\documentclass{article}
\usepackage{graphicx}
\usepackage[numbers,compress]{natbib}
\usepackage{authblk}
\usepackage{longtable}

\usepackage[utf8]{inputenc} 
\usepackage[T1]{fontenc}    
\usepackage{hyperref}       
\usepackage{url}            

\usepackage{amsfonts}       
\usepackage{nicefrac}       
\usepackage{microtype}      
\usepackage{subcaption}
\usepackage{multicol}
\usepackage[inline]{enumitem}
\usepackage{amsmath}
\usepackage{mathrsfs}
\usepackage{todonotes}

\DeclareMathOperator*{\argmin}{arg\,min}
\newcommand\norm[1]{\left \Vert #1 \right \Vert}
\newcommand\braces[1]{\left \lbrace #1 \right \rbrace}

\newcommand\func[2]{#1 \left( #2 \right)}
\newcommand\inv[1]{#1^{-1}}
\usepackage{algorithm2e}
\usepackage{footnote}
\usepackage{bbm}
\usepackage{arydshln}
\usepackage[title]{appendix}

\usepackage{rotating}
\newcommand*\rot{\rotatebox{90}}

\title{SpArSe: Sparse Architecture Search for CNNs on Resource-Constrained Microcontrollers}

\author[1]{Igor Fedorov \thanks{Corresponding author: igor.fedorov@arm.com}}
\affil[1]{ARM ML Research}
\author[2]{Ryan P. Adams}
\affil[2]{Princeton University}
\author[1]{Matthew Mattina}
\author[1]{Paul N. Whatmough}
\date{}

\begin{document}

\maketitle

\begin{abstract}
The vast majority of processors in the world are actually microcontroller units (MCUs), which find widespread use performing simple control tasks in applications ranging from automobiles to medical devices and office equipment.
The Internet of Things (IoT) promises to inject machine learning into many of these every-day objects via tiny, cheap MCUs.
However, these resource-impoverished hardware platforms severely limit the complexity of machine learning models that can be deployed. 
For example, although convolutional neural networks (CNNs) achieve state-of-the-art results on many visual recognition tasks, CNN inference on MCUs is challenging due to severe finite memory limitations.
To circumvent the memory challenge associated with CNNs, various alternatives have been proposed that do fit within the memory budget of an MCU, albeit at the cost of prediction accuracy.
This paper challenges the idea that CNNs are not suitable for deployment on MCUs. 
We demonstrate that it is possible to automatically design CNNs which generalize well, while also being small enough to fit onto memory-limited MCUs.
Our Sparse Architecture Search method combines neural architecture search with pruning in a single, unified approach, which learns superior models on four popular IoT datasets. 
The CNNs we find are more accurate and up to $4.35\times$ smaller than previous approaches, while meeting the strict MCU working memory constraint.
\end{abstract}

\section{Introduction}
\label{sec:introduction}
The microcontroller unit (MCU) is a truly ubiquitous computer.
MCUs are self-contained single-chip processors which are small $(\sim 1\text{cm}^2)$, cheap $(\sim \$1)$, and power efficient $(\sim 1 \text{mW})$. Applications are extremely broad, but often include seemingly banal tasks such as simple control and sequencing operations for everyday devices like washing machines, microwave ovens, and telephones.
The key advantage of MCUs over application specific integrated circuits (ASICs) is that they are programmed with software and can be readily updated to fix bugs, change functionality, or add new features.
The short time to market and flexibility of software has led to the staggering popularity of MCUs.
In the developed world, a typical home is likely to have around four general-purpose microprocessors.
In contrast, the number of MCUs is around three dozen~\citep{wikipedia2019microcontrollers}. 
A typical mid-range car may have about 30 MCUs.
The best public market estimates suggest that around 50 billion MCU chips will ship in 2019~\citep{embeddedmcumarket}, which far eclipses other computer chips like graphics processing units (GPUs), whose shipments totalled roughly 100 million units in 2018~\citep{gpu_volume}. 

MCUs can be highly resource constrained; Table~\ref{table:hardware} compares MCUs with bigger processors. 
The broad proliferation of MCUs relative to desktop GPUs and CPUs stems from the fact that they are orders of magnitude cheaper ($\sim 600\times$) and less power hungry ($\sim 250,000\times$). 
In recent years, MCUs have been used to inject intelligence and connectivity into everything from industrial monitoring sensors to consumer devices, a trend commonly referred to as the Internet of Things (IoT)~\citep{atzori2010internet,gubbi2013internet,meunier2014internet}. 
Deploying machine learning (ML) models on MCUs is a critical part of many IoT applications, enabling local autonomous intelligence rather than relying on expensive and insecure communication with the cloud~\citep{warden2019mcus}. 
In the context of supervised visual tasks, state-of-the-art (SOTA) ML models typically take the form of convolutional neural networks (CNNs) \citep{krizhevsky2012imagenet}. 
While tools for deploying CNNs on MCUs have started to appear \citep{utensor,tflitemicro,msrell}, the CNNs themselves remain far too large for the memory-limited MCUs commonly used in IoT devices. In the remainder of this work, we use MCU to refer specifically to IoT-sized MCUs, like the Micro:Bit. In contrast to this work, the majority of preceding research on compute/memory efficient CNN inference has targeted CPUs and GPUs~\citep{han2016eie,cai2018proxylessnas,yang2017designing,Yang_2018_ECCV,molchanov2016pruning,tan2018mnasnet,sandler2018mobilenetv2}.

To illustrate the challenge of deploying CNNs on MCUs, consider the seemingly simple task of deploying the well-known LeNet CNN on an Arduino Uno~\citep{arduinouno} to perform MNIST character recognition~\citep{lecun1998gradient}. 
Assuming the weights can be quantized to 8-bit integers, $420$ KB of memory is required to store the model parameters, which exceeds the Uno's $32$ KB of read-only (flash) memory. An additional $177$ KB of random access memory (RAM) is then required to store the intermediate feature maps produced by LeNet, which far exceeds the Uno's $2$ KB RAM. 
The dispiriting implication is that it is not possible to perform LeNet inference on the Uno. This has led many to conclude that CNNs should be abandoned on constrained MCUs~\citep{kumar2017resource,gupta2017protonn}.
Nevertheless, the sheer popularity of MCUs coupled with the dearth of techniques for leveraging CNNs on MCUs motivates our work, where we take a step towards bridging this gap.


\begin{table}
\caption{Processors for ML inference: estimated characteristics to indicate the relative capabilities.}
\vspace{3pt}
\label{table:hardware}
\centering
\resizebox{\linewidth}{!}{%
\begin{tabular}{l c c c c c}
        \hline
Processor                & Usecase   & Compute   & Memory & Power & Cost  \\ \hline


Nvidia 1080Ti GPU \citep{geforce}
                        & Desktop     & 10 TFLOPs/Sec  & 11 GB   & 250 W & \$700  \\

Intel i9-9900K CPU
\citep{intel_wikipedia,intel_linpack_blog}
                        & Desktop     & 500 GFLOPs/Sec      & 256 GB & 95 W & \$499  \\

Google Pixel 1 (Arm CPU) \citep{pixel} & Mobile    & 50 GOPs/Sec        & 4 GB & $\sim 5$ W & --  \\

Raspberry Pi (Arm CPU) \citep{raspberrypi}  & Hobbyist  & 50 GOPs/Sec        & 1 GB & 1.5 W & --  \\


Micro:Bit (Arm MCU) \citep{microbit}
                    & IoT       & 16 MOPs/Sec & 16 KB & $\sim 1$ mW & \$1.75  \\ 

{Arduino Uno (Microchip MCU)} 
\citep{arduinouno}
                        & IoT & 4 MOPs/Sec      & 2 KB & $\sim 1$ mW & \$1.14 \\ \hline
\end{tabular}}
\vspace{-1em}
\end{table}


Deployment of CNNs on MCUs is challenging along multiple dimensions, including power consumption and latency, but as the example above illustrates, it is the hard memory constraints that most directly prohibit the use of these networks. 
MCUs typically include two types of memory.
The first is static RAM, which is relatively fast, but volatile and small in capacity.
RAM is used to store intermediate data.
The second is flash memory, which is non-volatile and larger than RAM; it is typically used to store the program binary and any constant data.
Flash memory has very limited write endurance, and is therefore treated as read-only memory (ROM).
The two MCU memory types introduce the following constraints on CNN model architecture:
\begin{enumerate}[label=\textbf{C\arabic*}]
    \item\label{D:1}: The maximum size of intermediate feature maps cannot exceed the RAM capacity.
    \item\label{D:2}: The model parameters must not exceed the ROM (flash memory) capacity.
\end{enumerate}
To the best of our knowledge, there are currently no CNN architectures or training procedures that produce CNNs satisfying these MCU memory constraints~\citep{kumar2017resource,gupta2017protonn}.
This is true even ignoring the memory required for the runtime (in RAM) and the program itself (in ROM). 
The severe memory constraints for inference on MCUs have pushed research away from CNNs and toward simpler classifiers based on decision trees and nearest neighbors~\citep{kumar2017resource,gupta2017protonn}. We demonstrate for the first time that it is possible to design CNNs that are at least as accurate as \citet{kumar2017resource,gupta2017protonn} and at the same time satisfy \ref{D:1}-\ref{D:2}, even for devices with just $2$ KB of RAM. 
We achieve this result by designing CNNs that are heavily specialized for deployment on MCUs using a method we call \emph{Sparse Architecture Search} (SpArSe).
The key insight from SpArSe, is that combining neural architecture search (NAS) and network pruning allows us to balance generalization performance against tight memory constraints~\ref{D:1}-\ref{D:2}.
Critically, we enable SpArSe to search over pruning strategies in conjunction with conventional hyperparameters around morphology and training. 
Pruning enables SpArSe to quickly evaluate many sub-networks of a given network, thereby expanding the scope of the overall search. While previous NAS approaches have automated the discovery of performant models with reduced parameterizations, we are the first to simultaneously consider performance, parameter memory constraints, and inference-time working memory constraints.

\begin{figure}
    \centering
    \begin{subfigure}[t]{0.5\textwidth}
        \centering
        \includegraphics[width=\columnwidth]{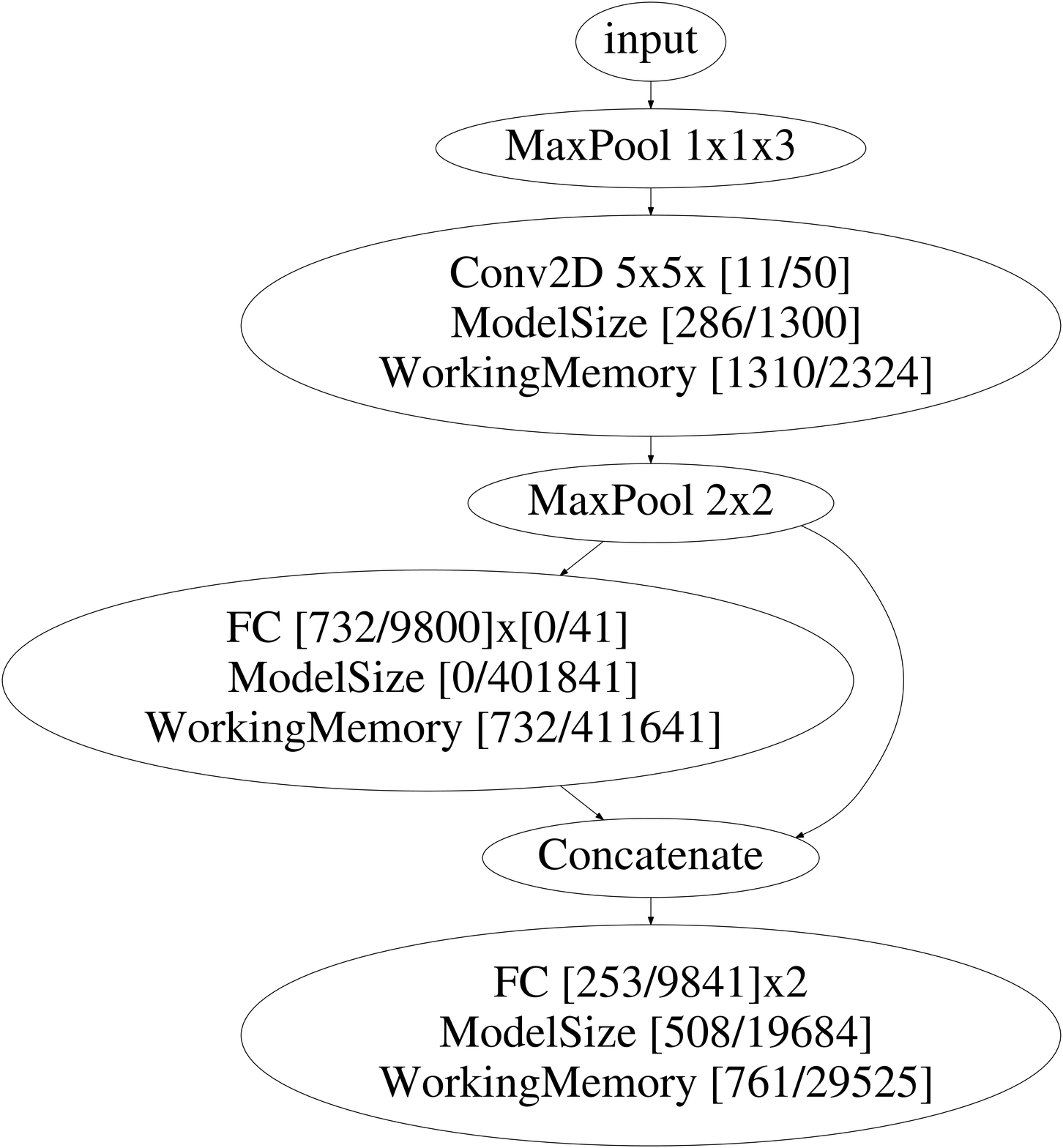}
        \caption{Acc = $73.84\%$, $\textsc{MS} = 1.31$ KB, \textsc{WM} = $1.28$ KB}
        \label{fig:comparison_a}
    \end{subfigure}%
    ~ 
    \begin{subfigure}[t]{0.5\textwidth}
        \centering
         \includegraphics[width=\columnwidth]{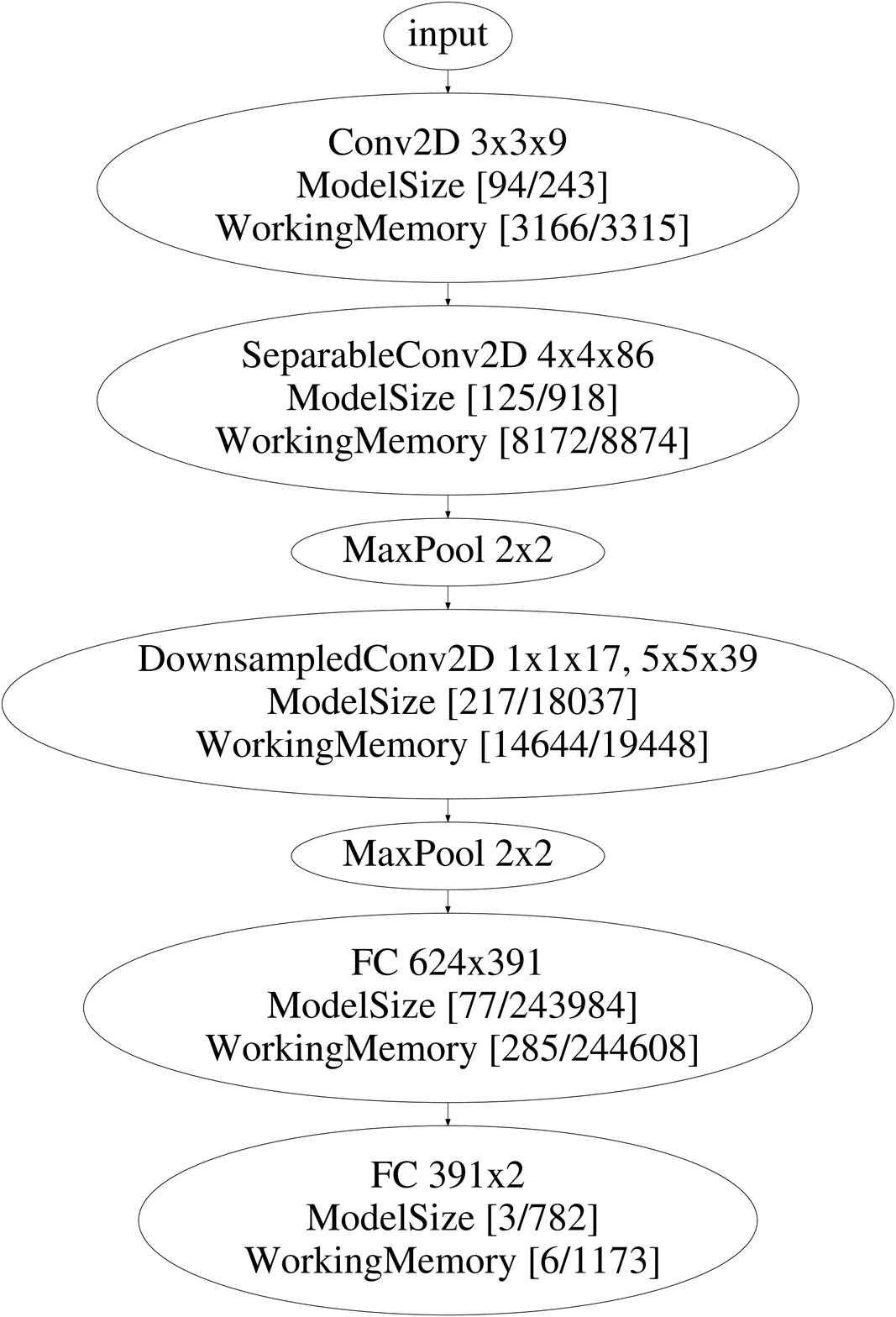}
        \caption{Acc=$73.58\%$, \textsc{MS} = $0.61$ KB, \textsc{WM} = $14.3$ KB}
         \label{fig:comparison_b}
    \end{subfigure}
    \caption{
    Model architectures found with best test accuracy on CIFAR10-binary, while optimizing for 
    (a) 2KB for both $\textsc{ModelSize}$ (MS) and $\textsc{WorkingMemory}$ (WM), and (b) minimum MS.
    Each node in the graph is annotated with MS and WS, and the values in square brackets show the quantities before and after pruning, respectively. 
    Optimizing for WM leads to a model that yields more than $11.2$x WM reduction. 
    Note that pruning has a considerable impact on the CNN.}
    \label{fig:comparison}
\end{figure}

We use SpArSe to uncover SOTA models on four datasets, in terms of accuracy and model size, outperforming both pruning of popular architectures and MCU-specific models \citep{kumar2017resource,gupta2017protonn}.
The multi-objective approach of SpArSe leads to new insights in the design of memory-constrained architectures.
Fig. \ref{fig:comparison_a} shows an example of a discovered architecture which has high accuracy, small model size, and fits within 2KB RAM.
By contrast, we find that optimizing networks solely to minimize the number of parameters  (as is typically done in the NAS literature, e.g., \citep{elsken2018efficient}), is not sufficient to identify networks that minimize RAM usage. Fig. \ref{fig:comparison_b} illustrates one such example.

\section{Related work}
CNNs designed for resource constrained inference have been widely published in recent years \citep{sandler2018mobilenetv2,howard2017mobilenets,zhang2018shufflenet}, motivated by the goal of enabling inference on mobile phone platforms. Advances include depth-wise separable layers \citep{sifre2014rigid}, deployment-centric pruning \citep{Yang_2018_ECCV,molchanov2016pruning}, and quantization techniques \citep{wang2018haq}. More recently, NAS has been leveraged to achieve even more efficient networks on mobile phone platforms \citep{cai2018proxylessnas,stamoulis2019singlepath}.

Although mobile phones are more constrained than general-purpose CPUs and GPUs, they still have many orders of magnitude more memory capacity and compute performance than MCUs (Table~\ref{table:hardware}).
In contrast, little attention has been paid to running CNNs on MCUs, which represent the most numerous compute platform in the world.
\citet{kumar2017resource} propose Bonsai, a pruned shallow decision tree with non-axis aligned decision boundaries. 
\citet{gupta2017protonn} propose a compressed k-nearest neighbors (kNN) approach (ProtoNN), where model size is reduced by projecting data into a low-dimensional space, maintaining a subset of prototypes to classify against, and pruning parameters. We build upon \citet{kumar2017resource,gupta2017protonn} by targeting the same MCUs, but using NAS to find CNNs which are at least as small and more accurate.

Algorithms for identifying performant CNN architectures have received significant attention recently \citep{zoph2016neural,elsken2018efficient,cai2018proxylessnas,liu2018darts,guo2019single,elsken2019neural,automl_overview}.
The approaches closest to SpArSe are \citet{stamoulis2019singlepath,elsken2018efficient}. 
In \citet{stamoulis2019singlepath}, the authors optimize the kernel size and number of feature maps of the MBConv layers in a MobileNetV2 backbone \citep{sandler2018mobilenetv2} by expressing each of the layer choices as a pruned version of a superkernel. In some ways, \citet{stamoulis2019singlepath} is less a NAS algorithm and more of a structured pruning approach, given that the only allowed architectures are reductions of MobileNetV2. SpArSe does not constrain architectures to be pruned versions of a baseline, which can be too restrictive of an assumption for ultra small CNNs. SpArSe is not based on an existing backbone, giving it greater flexibility to extend to different problems. Like \citet{elsken2018efficient}, SpArSe uses a form of weight sharing called network morphism \citep{wei2016network} to search over architectures without training each one from scratch. SpArSe extends the concept of morphisms to expedite training and pruning CNNs. Because
\citet{elsken2018efficient} seek compact architectures by using the number of network edges as one of the objectives in the search, potential gains from weight sparsity are ignored, which can be significant (Section \ref{sec:nas_with_pruning_discussion} \citep{frankle2018lottery,gale2019state}). Moreover, since SpArSe optimizes both the architecture and weight sparsity, \citet{elsken2018efficient} can be seen as a special case of SpArSe.



\section{SpArSe framework: CNN design as \\ multi-objective optimization}


Our approach to designing a small but performant CNN is to specify a multi-objective optimization problem that balances the competing criteria.
We denote a point in the design space as~${\Omega = \braces{\alpha,\vartheta,\omega, \theta}}$, in which: ${\alpha = \braces{V,E}}$ is a directed acyclic graph describing the network connectivity, where $V$ and $E$ denote the set of graph vertices and edges;~$\omega$~denotes the network weights;~$\vartheta$ represents the operations performed at each edge, i.e. convolution, pooling, etc.; and $\theta$ are hyperparameters governing the training process. 
The vertices~${v_i,v_j \in V}$ represent network neurons, which are connected to each other if~${\func{}{v_i,v_j} \in E}$ through an operation $\vartheta_{ij}$ parameterized by $\omega$. The competing objectives in the present work of targeting constrained MCUs are:
\begin{align}
    f_1(\Omega) &= 1 - \textsc{ValidationAccuracy}(\Omega) \label{eq:f1} \\
    f_2(\Omega) &= \textsc{ModelSize}(\omega) \label{eq:f2} \\
    f_3(\Omega) &= \max_{l \in 1,\ldots,L} \textsc{WorkingMemory}_l(\Omega) \label{eq:f3}
\end{align}
where $\textsc{ValidationAccuracy}(\Omega)$ is the accuracy of the trained model on validation data, $\textsc{ModelSize}(\omega)$, or MS, is the number of bits needed to store the model parameters~$\omega$, $\textsc{WorkingMemory}_l(\Omega)$ is the working memory in bits needed to compute the output of layer~$l$, with the maximum taken over the~$L$ layers to account for in-place operations.
We refer to \eqref{eq:f3} as the working memory (WM) for $\Omega$.

There is no single $\Omega$ which minimizes all of $\eqref{eq:f1}-\eqref{eq:f3}$ simultaneously. 
For instance, \eqref{eq:f1} prefers large networks with many non-zero weights whereas \eqref{eq:f2} favors networks with no weights. 
Likewise, \eqref{eq:f3} prefers configurations with small intermediate representations, whereas \eqref{eq:f2} has no preference as to the size of the feature maps. 
Therefore, in the context of CNN design, it is more appropriate to seek the set of Pareto optimal configurations, where $\Omega^\star$ is Pareto optimal if
\begin{align*}
f_k(\Omega^\star) \leq f_k(\Omega) \; \forall k,\Omega \qquad \text{and} \qquad \exists j: f_j(\Omega^\star) < f_j(\Omega) \;  \forall \Omega \neq \Omega^\star\,.
\end{align*}
The concept of Pareto optimality is appealing for multi-objective optimization, as it allows the ready identification of optimal designs subject to arbitrary constraints in a subset of the objectives.

\subsection{Search space}
Our search space is designed to encompass CNNs of varying depth, width, and connectivity. Each graph consists of optional input downsampling followed by a variable number of blocks, where each block contains a variable number of convolutional layers, each parametrized by its own kernel size, number of output channels, convolution type, and padding. 
We consider regular, depthwise separable, and downsampled convolutions, where we define a downsampled convolution to be a $1 \times 1$ convolution that downsamples the input in depth, followed by a regular convolution. Each convolution is followed by optional batch-normalization, ReLU, and spatial downsampling through pooling of a variable window size. Each set of two consecutive convolutions has an optional residual connection. Inspired by the decision tree approach in \citet{kumar2017resource}, we let the output layer use features at multiple scales by optionally routing the output of each block to the output layer through a fully connected (FC) layer (see Fig. \ref{fig:comparison_a}). All of the FC layer outputs are merged before going through an FC layer that generates the output. The search space also includes parameters governing CNN training and pruning. The Appendix contains a complete description of the search space.

\subsection{Quantifying memory requirements}
The~$\textsc{ValidationAccuracy}(\Omega)$ metric is readily available for trained models via a held-out validation set or by cross-validation.
However, the memory constraints of interest in this work demand more careful specification.
For simplicity, we estimate the model size as 
\begin{align}
    \textsc{ModelSize}(\omega) &\approx \norm{\omega}_0 \label{eq:size}.
\end{align}
For working memory, we consider two different models
:
\begin{align}
    \textsc{WorkingMemory}_l^1(\Omega) &\approx \norm{x_l}_0 + \norm{\omega_l}_0 \label{eq:workingmem} \\
    \textsc{WorkingMemory}_l^2(\Omega) &\approx \norm{x_l}_0 + \norm{y_l}_0 \label{eq:workingmem2} 
\end{align}
where $x_l$, $y_l$, and $\omega_l$ are the input, output, and weights for layer $l$, respectively. The assumption in \eqref{eq:workingmem} is that the inputs to layer $l$ and the weights need to reside in RAM to compute the output, which is consistent with deployment tools like \citep{utensor} which allow layer outputs to be written to an SD card. The model in \eqref{eq:workingmem2} is also a standard RAM usage model, adopted in \citep{visual_wake}, for example. For merge nodes that sum two vector inputs $x_l^1$ and $x_l^2$, we set $x_l = \begin{bmatrix} \left(x_l^1\right)^T & \left(x_l^2\right)^T  \end{bmatrix}^T$ in \eqref{eq:workingmem}-\eqref{eq:workingmem2}.
The reliance of \eqref{eq:size}-\eqref{eq:workingmem2} on the $\ell_0$ norm is motivated by our use of pruning to minimize the number of non-zeros in both $\omega$ and $\braces{x_l}_{l=1}^L$, which is also the compression mechanism used in related work~\citep{kumar2017resource,gupta2017protonn}. 
Note that \eqref{eq:size}-\eqref{eq:workingmem2} are reductive to varying degrees.
However, since SpArSe is a black-box optimizer, the measures in \eqref{eq:size}-\eqref{eq:workingmem2} can be readily updated as MCU deployment toolchains mature.

\subsection{Neural network pruning}\label{sec:pruning}
Pruning \citep{lecun1990optimal,molchanov2017variational,Carreira_2018_CVPR,louizos2017bayesian} is essential to MCU deployment using SpArSe, as it heavily reduces the model size \eqref{eq:size} and working memory \eqref{eq:workingmem}/\eqref{eq:workingmem2} without significantly impacting classification accuracy. Pruning is a procedure for zeroing out network parameters $\omega$ and can be seen as a way to generate a new set of parameters~$\bar{\omega}$ that have lower~$\norm{\bar{\omega}}_0$. We consider both unstructured and structured, or channel \citep{he2017channel}, pruning, where the difference is that the latter prunes away entire groups of weights corresponding to output feature maps for convolution layers and input neurons for FC layers. Both forms of pruning reduce $\norm{\omega}_0$ and, consequently, \eqref{eq:size}-\eqref{eq:workingmem}. Structured pruning is critical for reducing \eqref{eq:workingmem}-\eqref{eq:workingmem2} because it provides a mechanism for reducing the size of layer inputs. We use Sparse Variational Dropout (SpVD) \citep{molchanov2017variational} and Bayesian Compression (BC) \citep{louizos2017bayesian} to realize unstructured and structured pruning, respectively. Both approaches assume a sparsity promoting prior on the weights and approximate the weight posterior by a distribution parameterized by $\phi$. See the Appendix for a description of SpVD and BC. Notably, $\phi$ contains all of the information about the network weight values as well as which weights to prune.





\subsection{Multi-objective Bayesian optimization}\label{sec:MOBO}
SpArSe consists of three stages, where each stage~$m$ samples~$T_m$ configurations. At iteration $n$, a new configuration~$\Omega^n$ is generated by the multi-objective Bayesian optimizer (MOBO) with probability~$\rho_m$ and uniformly at random with probability~${1-\rho_m}$. We adopt the combination of model-based and entirely random sampling from \citep{falkner2018bohb} to increase search space coverage. 
The optimizer considers candidates which are morphs of previous configurations and returns both the new and reference configurations (Section \ref{sec:morphism}). The parameters of the new architecture are then inherited from the reference before being retrained and pruned.

SpArSe uses a MOBO based on the idea of random scalarizations \citep{paria2018flexible}. 
The MOBO approach is appealing as it builds flexible nonparametric models of the unknown objectives and enables reasoning about uncertainty in the search for the Pareto frontier.
A scalarized objective is given by
\begin{align}\label{eq:random scalarization}
    \func{g}{\braces{\lambda_k}_{k=1}^K,\Omega} = \max_{k \in 1,\ldots,K} \lambda_k f_k(\Omega)
\end{align}
where $\lambda_k$ is drawn randomly. Choosing the domain of the prior on $\lambda_k$ allows the user to specify preferences about the region of the Pareto frontier to explore. 
For example, IoT practitioners may care about models with less than $1000$ parameters. 
Since $f_k(\Omega)$ is unknown in practice, it is modeled by a Gaussian process \citep{rasmussen2003gaussian} with a kernel $\func{\kappa}{\cdot,\cdot}$ that supports the types of variables included in $\Omega$, i.e., real-valued, discrete, categorical, and hierarchically related variables \citep{swersky2014raiders,garrido2018dealing}. 
A new $\Omega^n$ is sampled by minimizing \eqref{eq:random scalarization} through Thompson sampling. This MOBO yields better coverage of the Pareto frontier than the deterministic scalarization methods used in \citep{cai2018proxylessnas,stamoulis2019singlepath}.

\subsection{Network morphism}
\label{sec:morphism}
Evaluating each configuration $\Omega^n$ from a random initialization is slow, as evidenced by early NAS works which required thousands of GPU days~\citep{zoph2016neural,zoph2018learning}. 
Search time can be reduced by constraining each proposal to be a morph of a reference~${\Omega^r \in \braces{\Omega^j}_{j=0}^{n-1}}$ \citep{elsken2018efficient}. Loosely speaking, we say that~$\Omega^n$ is a morph of~$\Omega^r$ if most of the elements in~$\Omega^n$ are identical to those in~$\Omega^r$. 
The advantage of using morphism to generate~$\Omega^n$ is that most of~$\phi^n$ can be inherited from~$\phi^r$, where~$\phi^r$ denotes the weight posterior parameters for configuration~$\Omega^r$. 
Initializing~$\phi^n$ in this way means that~$\Omega^n$ inherits knowledge about the value and pruning mask for most of its weights.
Compared to running SpVD/BC from scratch, morphisms enable pruning proposals using $2$-$8\times$ fewer epochs, depending on the dataset. Further details on morphism are given in the Appendix, including allowed morphs.

Because our search space includes such a diversity of parameters, including architectural parameters, pruning hyperparameters, etc., we find it helpful to perform the search in stages, where each successive stage increasingly limits the set of possible proposals. This coarse-to-fine search enables exploring decisions at increasing granularity, to wit: Stage 1) A candidate configuration can be generated by applying modifications to any of $\braces{\Omega^r}_{r=1}^{n-1}$, Stage 2) The allowable morphs are restricted to the pruning parameters, Stage 3) The reference configurations are restricted to the Pareto optimal points.

\section{Results}
\vspace{-1em}
We report results on a variety of datasets: MNIST $(55e3,5e3,10e3)$ \citep{lecun1998gradient}, CIFAR10 $(45e3,5e3,10e3)$ \citep{krizhevsky2009learning}, CUReT $(3704,500,1408)$ \citep{varma2005statistical}, and Chars4k $(3897,500,1886)$ \citep{de2009character}, corresponding to classification problems with $10$, $10$, $61$, and $62$ classes, respectively, with the training/validation/test set sizes provided in parentheses. To match the setup in \citep{kumar2017resource}, we also report on binary versions of these datasets, meaning that the classes are split into two groups and re-labeled. The only pre-processing we perform is mean subtraction and division by the standard deviation. Experiments were run on four NVIDIA RTX 2080 GPUs. We compare against previous SOTA works: Bonsai \citep{kumar2017resource}, ProtoNN \citep{gupta2017protonn}, Gradient Boosted Decision Tree Ensemble Pruning \citep{dekel2016pruning}, kNN, and radial basis function support vector machine (SVM). We do not compare against previous NAS works because they have not addressed the memory-constrained classification problem addressed here.
\begin{figure}
    \centering
    \begin{subfigure}[t]{0.5\textwidth}
        \centering
        \includegraphics[width=\textwidth]{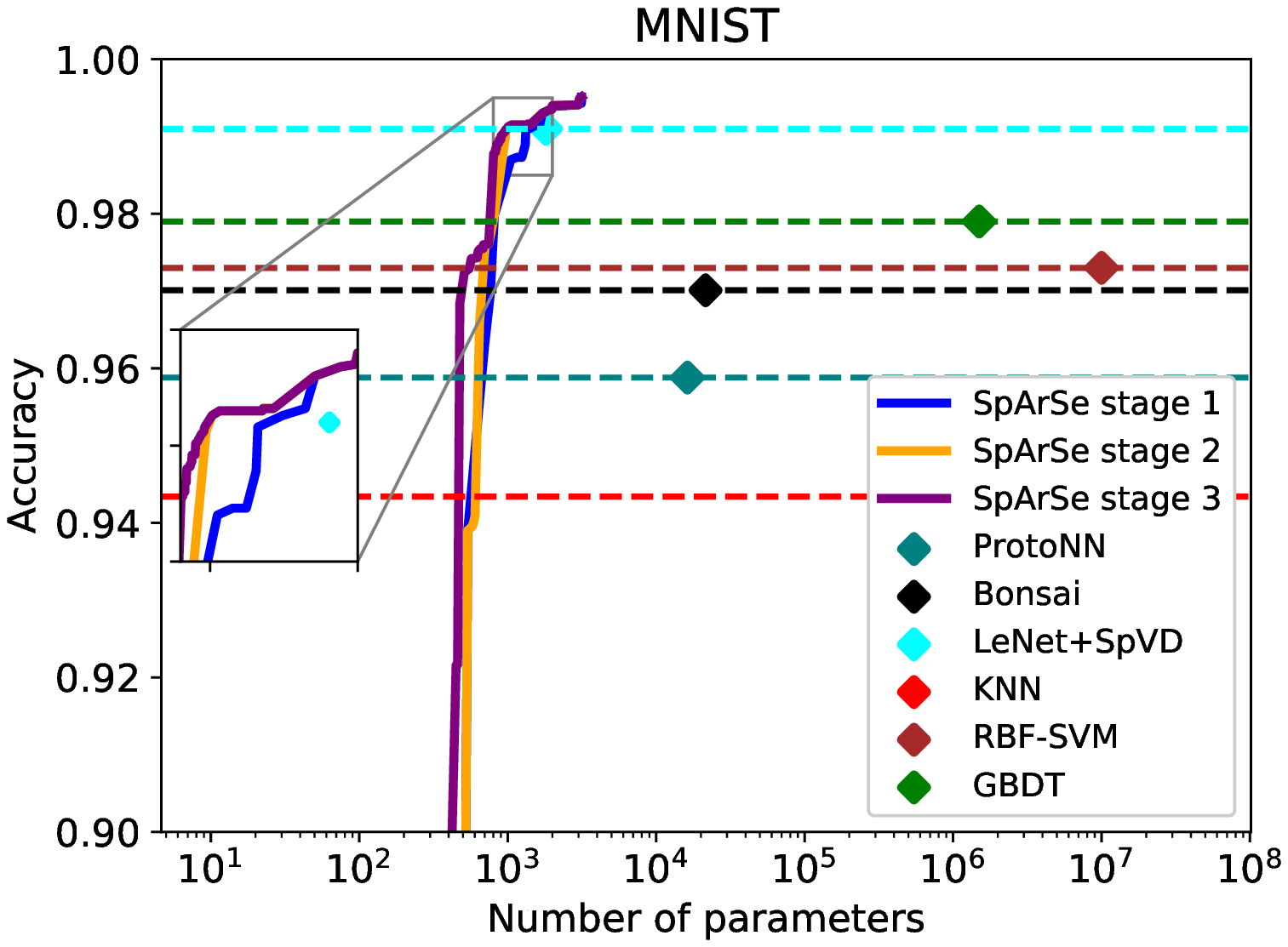}
        \vspace{-20pt}
        \caption{}
    \end{subfigure}%
    ~ 
    \begin{subfigure}[t]{0.5\textwidth}
        \centering
        \includegraphics[width=\textwidth]{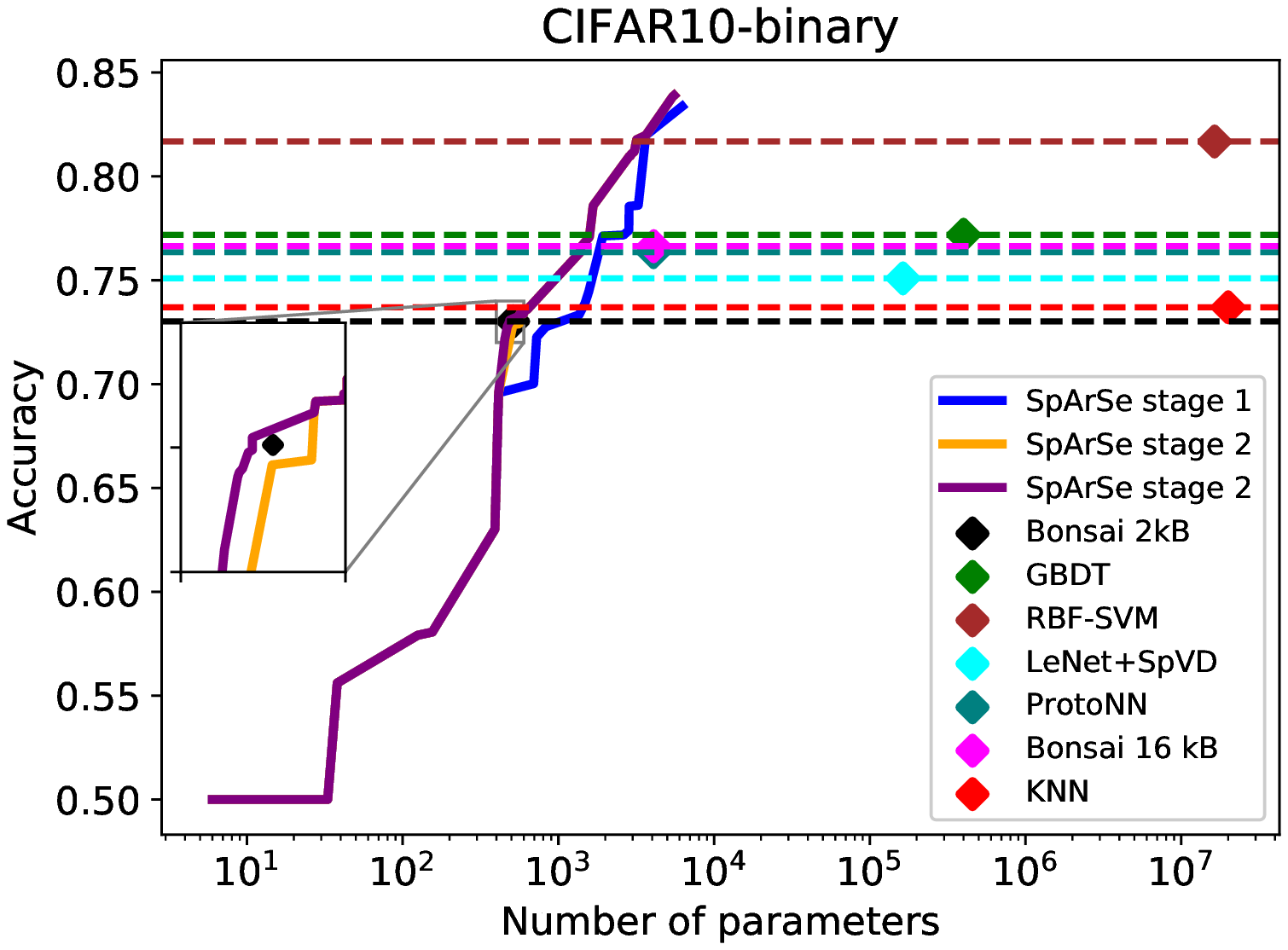}
        \vspace{-20pt}
        \caption{}
    \end{subfigure}
    ~
    \begin{subfigure}[t]{0.5\textwidth}
        \centering
        \includegraphics[width=\textwidth]{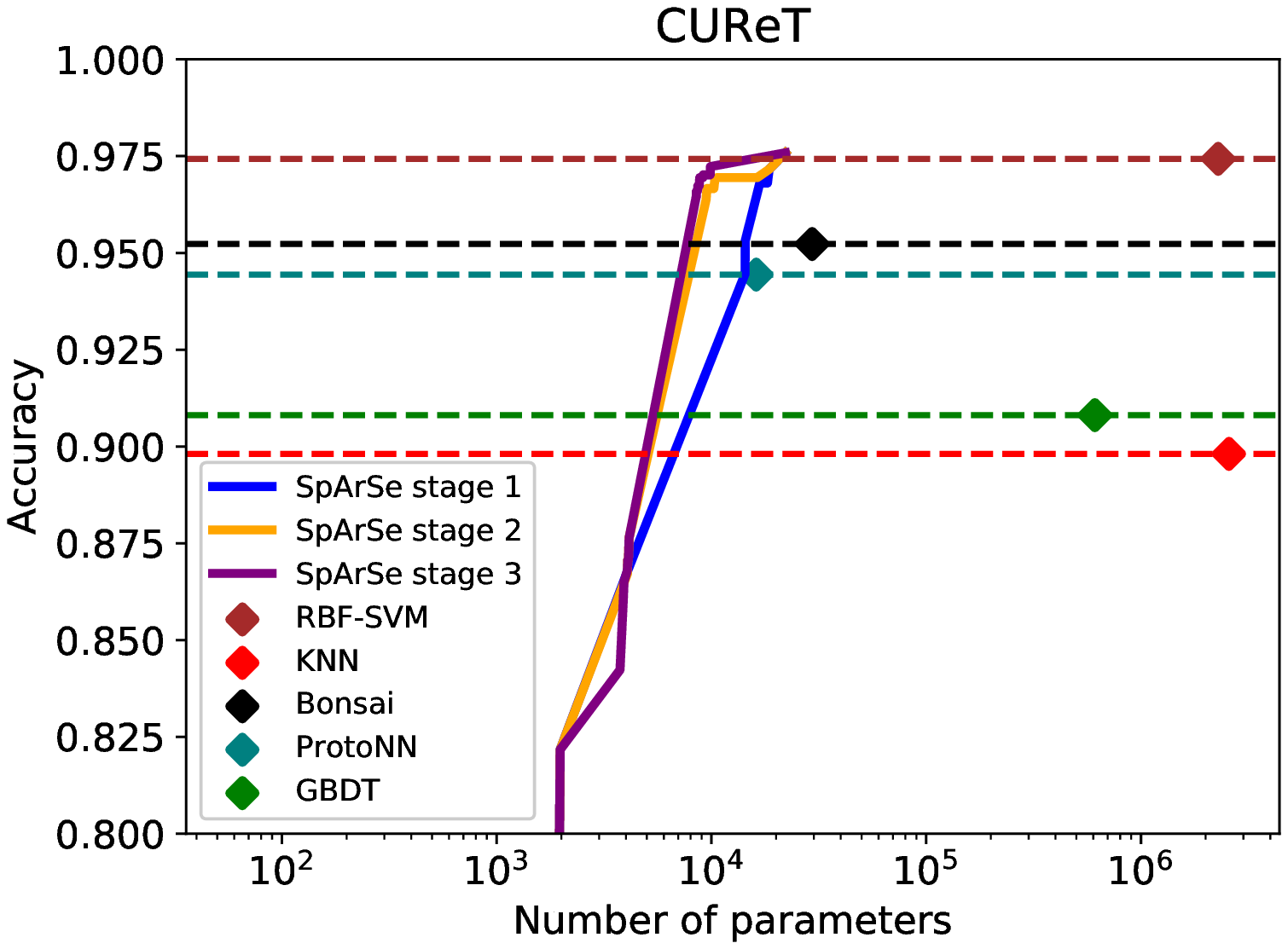}
        \vspace{-20pt}
        \caption{}
    \end{subfigure}%
    ~ 
    \begin{subfigure}[t]{0.5\textwidth}
        \centering
        \includegraphics[width=\textwidth]{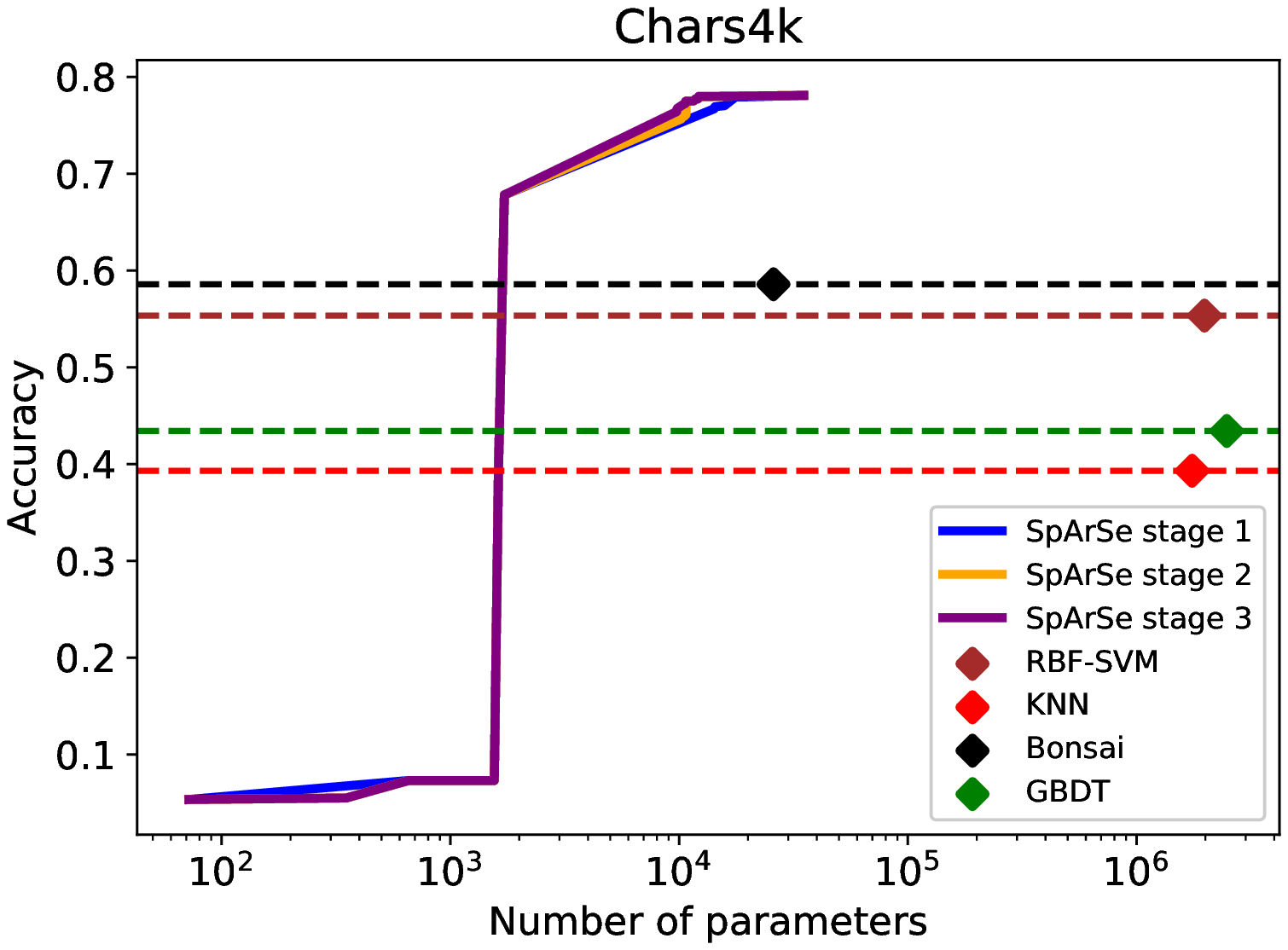}
        \vspace{-20pt}
        \caption{}
    \end{subfigure}
    \vspace{-8pt}
    \caption{SpArSe results from minimization of $\left(1 - \textsc{ValidationAccuracy}(\Omega)\right),\textsc{ModelSize}(\omega)$.}
    \label{fig:pareto acc params}
    \vspace{-1.5em}
\end{figure}

\vspace{-0.5em}
\subsection{Models optimized for number of parameters}\label{sec:parameter minimization}
First, we address \ref{D:2} by showing that SpArSe finds CNNs with higher accuracy and fewer parameters than previously published methods. We use unstructured pruning and optimize $\braces{\func{f_k}{{\Omega}}}_{k=1}^2$. 
Fig. \ref{fig:pareto acc params} shows the Pareto curves for SpArSe and confirms that it finds smaller and more accurate models on all datasets. For each competing method, we also report the SpArSe-obtained configuration which attains the same or higher test accuracy and minimum number of parameters, which we term the dominating configuration. 
Results are shown in Table \ref{table:dominating}. To confirm that SpArSe learns non-trivial solutions, we compare  with applying SpVD pruning to LeNet in Fig. \ref{fig:pareto acc params} and Table \ref{table:dominating}.

\vspace{-0.5em}
\subsection{Models optimized for total memory footprint}
\label{sec:results working memory}
Next, we demonstrate that SpArSe resolves \ref{D:1}-\ref{D:2} by finding CNNs that consume less device memory than Bonsai \citep{kumar2017resource}. We use structured pruning and optimize $\braces{\func{f_k}{{\Omega}}}_{k=1}^3$. We quantize weights and activations to one byte to yield realistic memory calculations and for fair comparison with Bonsai \citep{tf_fake_quant}. Table \ref{table:working memory} compares SpArSe to Bonsai in terms of accuracy, MS, and WM under the model in \eqref{eq:workingmem}. For all datasets and metrics, SpArSe yields CNNs which outperform Bonsai. For MNIST, Bonsai reports performance on a binarized dataset, whereas we use the original ten-class problem, i.e., we solve a significantly more complex problem with fewer resources. Table \ref{table:working memory accurate} reports results for WM model \eqref{eq:workingmem2}, showing that SpArSe outperforms Bonsai across all metrics on the MNIST, CUReT, and Chars4k datasets, whereas Bonsai achieves higher accuracy on CIFAR10.
For validation, we use uTensor~\citep{utensor} to convert CNNs from SpArSe into baremetal C++, which we compile using mbed-cli~\citep{mbed_cli} and deploy on the STM32~\citep{stm32} MCU.

\subsection{What SpArSe reveals about pruning}
\label{sec:nas_with_pruning_discussion}
\begin{figure}
    \centering
    \begin{subfigure}[t]{0.3\textwidth}
        \centering
        \includegraphics[width=\textwidth]{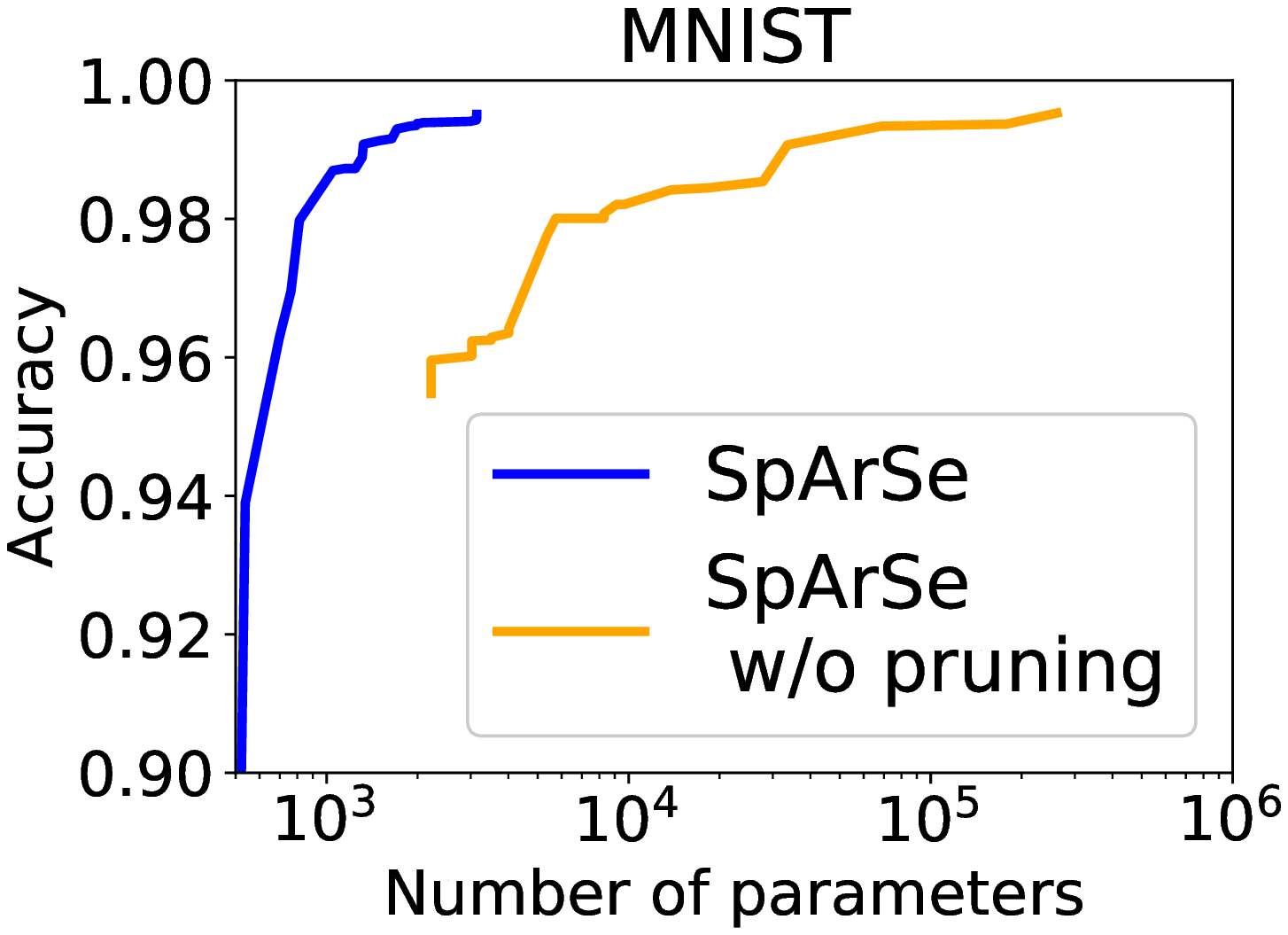}
        \caption{}
        \label{fig:ablation}
    \end{subfigure}
    ~
    \begin{subfigure}[t]{0.3\textwidth}
        \centering
        \includegraphics[width=\textwidth]{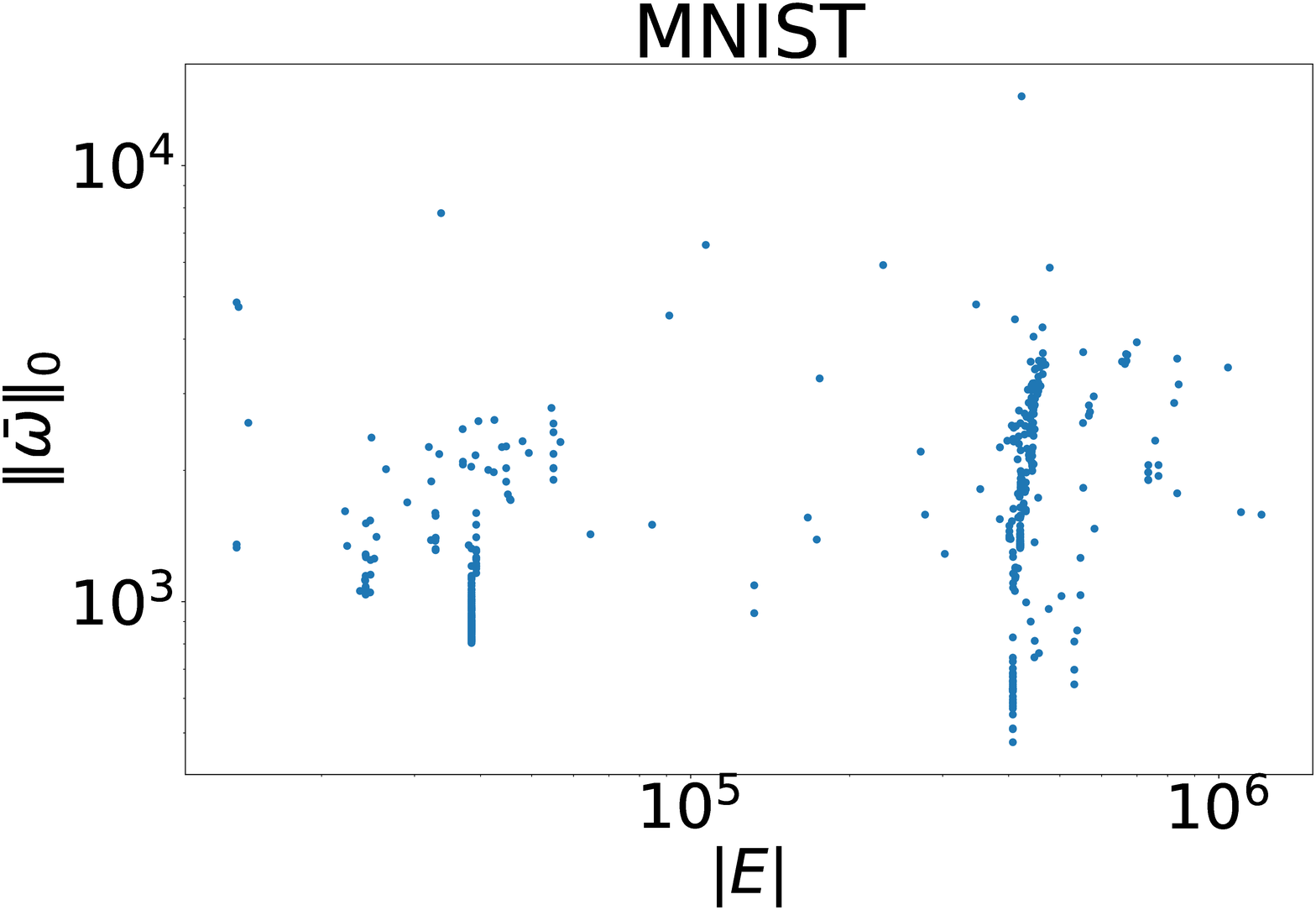}
        \caption{}
        \label{fig:mnist_pruned_vs_unpruned}
    \end{subfigure}%
    ~ 
    \begin{subfigure}[t]{0.3\textwidth}
        \centering
        \includegraphics[width=\textwidth]{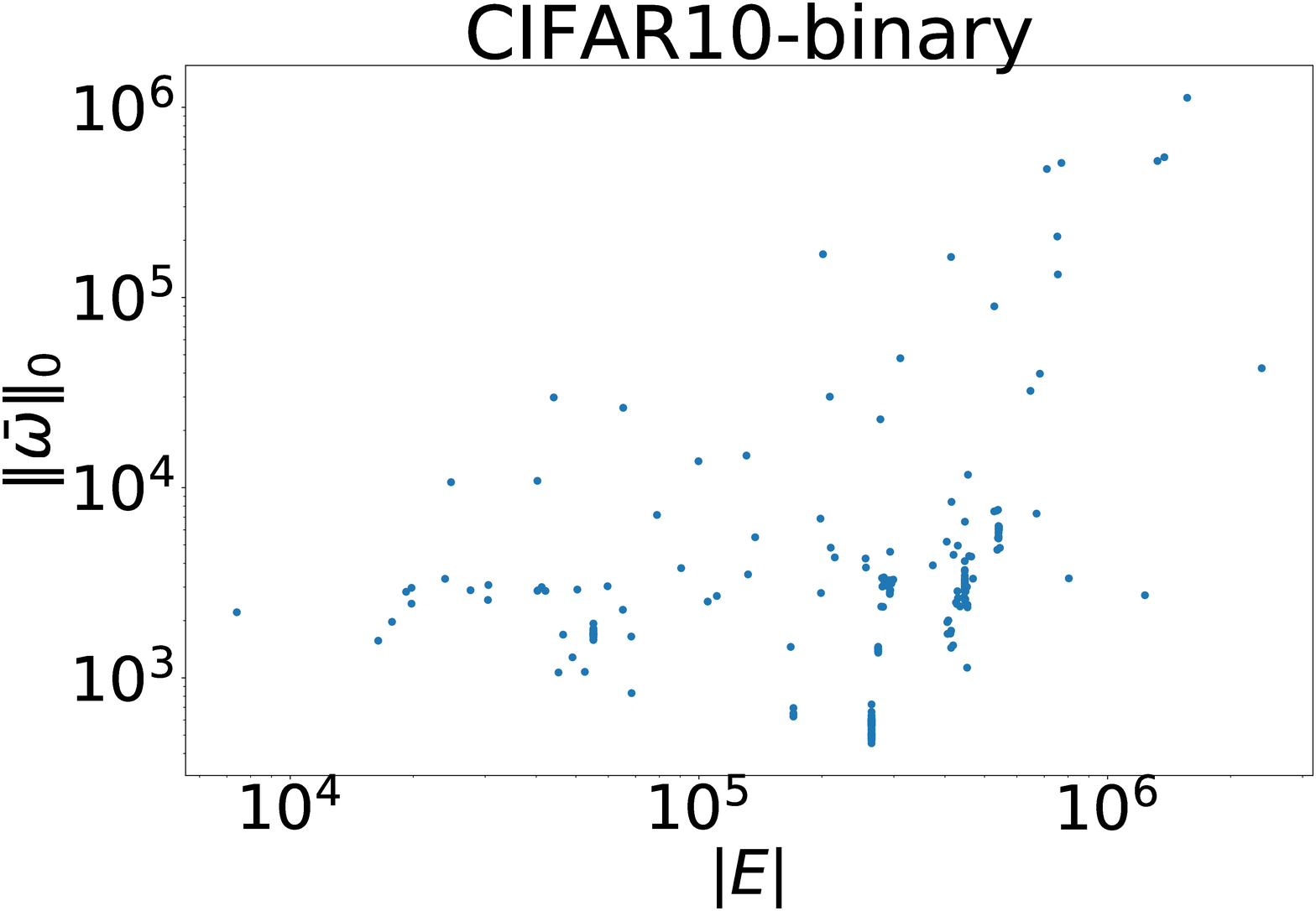}
        \caption{}
        \label{fig:cifar_pruned_vs_unpruned}
    \end{subfigure}
    \caption{Fig. \ref{fig:ablation} shows the Pareto frontier of SpArSe with and without pruning, where both experiments sample the same number (325) of configurations. Fig. \ref{fig:mnist_pruned_vs_unpruned}-\ref{fig:cifar_pruned_vs_unpruned} show scatter plots of $\vert E \vert$ versus $\norm{\bar{\omega}}_0$ for the best performing configurations from the experiment in Section \ref{sec:parameter minimization}. Fig. \ref{fig:mnist_pruned_vs_unpruned}: MNIST networks with $>95\%$ accuracy. Fig. \ref{fig:cifar_pruned_vs_unpruned}: CIFAR10-binary networks with $>70\%$ accuracy. }
    \label{fig:scatter}
\end{figure}

Pruning can be considered a form of NAS, where~$\bar{\omega}$ represents a sub-network of~$\braces{\alpha,\vartheta,\omega}$ given by~$\braces{\braces{V,E_p},\vartheta,\omega}$, and~${E_p \subseteq E}$ contains only the edges for which~$\bar{\omega}$ is non-zero \citep{frankle2018lottery}. The question then becomes, should one look for $E_p$ directly or begin with a large edge-set $E$ and prune it? There is conflicting evidence whether the same validation accuracy can be achieved by both approaches \citep{frankle2018lottery,gale2019state,liu2018rethinking}. Importantly, previous NAS approaches have focused on searching for $E_p$ directly by using $\vert E \vert$ as one of the optimization objectives \citep{elsken2018efficient}. On the other hand, SpArSe is able to explore both strategies and learn the optimal interaction between network graph $\alpha$, operations $\vartheta$, and pruning. Fig. \ref{fig:ablation} compares SpArSe to SpArSe without pruning on MNIST. The results show that including pruning as part of the optimization yields roughly a $80$x reduction in number of parameters, indicating that the formulation of SpArSe is better suited to designing tiny CNNs compared to \citep{elsken2018efficient}. To gain more insight, we show scatter plots of~${\vert E \vert}$ versus~${\norm{\bar{\omega}}_0}$ for the best-performing configurations on two datasets in Fig. \ref{fig:mnist_pruned_vs_unpruned}-\ref{fig:cifar_pruned_vs_unpruned}, revealing two important trends (see the Appendix for results on the Chars4k and CUReT datasets). First, $\norm{\bar{\omega}}_0$ tends to increase with increasing~$\vert E \vert$ for~${\vert E \vert}$ greater than some threshold $\zeta$. This suggests that optimizing $\vert E \vert$ can be a proxy for optimizing $\norm{\bar{\omega}}_0$ when targeting large networks. At the same time, $\norm{\bar{\omega}}_0$ tends to decrease with increasing $\vert E \vert$ for $\vert E \vert < \zeta$, which has implications for both NAS and pruning in the context of small CNNs. Fig. \ref{fig:mnist_pruned_vs_unpruned}-\ref{fig:cifar_pruned_vs_unpruned} suggest that $\vert E \vert$ is not always indicative of weight sparsity, such that minimizing $\vert E \vert$ would actually lead to ignoring graphs with more edges but the same amount of non-zero weights. Since CNNs with more edges contain more subgraphs, it is possible that one of these subgraphs actually has better accuracy and the same number of non-zero weights as the subgraphs of a graph with less edges. The key is that pruning provides a mechanism for uncovering such high performing subgraphs \citep{frankle2018lottery}.  
\begin{table}
  \caption{Dominating configurations for the parameter minimization experiment in Section \ref{sec:results working memory}. SpArSe models are listed on top and the competing method on bottom. SpArSe finds CNNs that are more accurate and have fewer parameters than competing methods. The amount of time spent obtaining each dominating configuration is reported in GPU days (GPUD).}
  \vspace{3pt} 
  \label{table:dominating}
  \centering
  \resizebox{\linewidth}{!}{%
  \begin{tabular}{cccc|ccc|ccc|ccc}
        \hline
        &  \multicolumn{3}{c}{MNIST} & \multicolumn{3}{c}{CIFAR10-binary} & \multicolumn{3}{c}{CUReT} & \multicolumn{3}{c}{Chars4k} \\
        & \rot{Acc} & \rot{$\norm{\omega}_0$} & \rot{GPUD} & \rot{Acc} & \rot{$\norm{\omega}_0$} & \rot{GPUD} & \rot{Acc} & \rot{$\norm{\omega}_0$} & \rot{GPUD} & \rot{Acc} & \rot{$\norm{\omega}_0$} & \rot{GPUD} \\ \hline
        Bonsai &  $\begin{array}{c} \textbf{97.24} \\ 97.01 \end{array}$ & $\begin{array}{c} \textbf{510}\\2.15e4 \end{array} $ & 11 & $\begin{array}{c} \textbf{73.08} \\ 73.02 \end{array}$ & $\begin{array}{c} \textbf{487} \\ 512 \end{array}$ & 1 & $\begin{array}{c} \textbf{96.45} \\ 95.23 \end{array}$ & $\begin{array}{c} \textbf{8.5e3} \\ 2.9e4 \end{array}$ & 1 & $\begin{array}{c} \textbf{67.82} \\ 58.59 \end{array}$ & $\begin{array}{c} \textbf{1.7e3} \\ 2.6e4 \end{array}$ & 1\\ \hdashline
        Bonsai (16 kB) & -- & -- & -- & $\begin{array}{c} \textbf{76.66} \\ 76.64 \end{array}$ & $\begin{array}{c} \textbf{1.4e3} \\ 4.1e3\end{array}$ & 9 & -- & -- & -- & -- & -- & -- \\ \hdashline
        ProtoNN & $\begin{array}{c} \textbf{96.84}\\ 95.88 \end{array}$ & $\begin{array}{c} \textbf{476} \\ 1.6e4 \end{array}$ & 11 & $\begin{array}{c} \textbf{76.56} \\ 76.35 \end{array}$ & $\begin{array}{c} \textbf{1.4e3} \\ 4.1e3\end{array}$ & 10 & $\begin{array}{c} \textbf{96.45} \\ 94.44 \end{array}$ & $\begin{array}{c} \textbf{8.5e3} \\ 1.6e4\end{array}$ & 1 & -- & -- & -- \\ \hdashline
        GBDT & $\begin{array}{c} \textbf{98.78}\\ 97.90 \end{array}$ & $\begin{array}{c} \textbf{804} \\ 1.5e6 \end{array}$ & 11 & $\begin{array}{c} \textbf{77.90} \\ 77.19 \end{array}$ & $\begin{array}{c} \textbf{1.6e3} \\ 4e5 \end{array}$ & 8 & $\begin{array}{c} \textbf{96.45} \\ 90.81 \end{array}$ & $\begin{array}{c} \textbf{8.5e3} \\ 6.1e5 \end{array}$ & 1 &  $\begin{array}{c} \textbf{67.82} \\ 43.34 \end{array}$ & $\begin{array}{c} \textbf{1.7e3} \\ 2.5e6 \end{array}$ & 1 \\ \hdashline
        kNN & $\begin{array}{c} \textbf{96.84} \\ 94.34 \end{array}$ & $\begin{array}{c} \textbf{476} \\ 4.71e7 \end{array}$ & 11 & $\begin{array}{c} \textbf{76.34} \\ 73.70 \end{array}$ & $\begin{array}{c} \textbf{1.4e3} \\ 2e7 \end{array}$ & 10 & $\begin{array}{c} \textbf{96.45} \\ 89.81 \end{array}$ & $\begin{array}{c} \textbf{8.5e3} \\ 2.6e6 \end{array}$ & 2 &  $\begin{array}{c} \textbf{67.82} \\ 39.32 \end{array}$ & $\begin{array}{c} \textbf{1.7e3} \\ 1.7e6 \end{array}$ & 1\\ \hdashline
        RBF-SVM & $\begin{array}{c} \textbf{97.42} \\ 97.30\end{array}$ & $\begin{array}{c} \textbf{569} \\ 1e7 \end{array}$ & 10 & $\begin{array}{c} \textbf{81.77} \\ 81.68 \end{array}$ & $\begin{array}{c} \textbf{3.2e3} \\ 1.6e7 \end{array}$ & 3 &  $\begin{array}{c} \textbf{97.58} \\ 97.43 \end{array}$ & $\begin{array}{c} \textbf{2.2e4} \\ 2.3e6 \end{array}$ & 2 & $\begin{array}{c} \textbf{67.82} \\ 48.04 \end{array}$ & $\begin{array}{c} \textbf{1.7e3} \\ 2e6 \end{array}$ & 1\\ \hdashline
        LeNet + SpVD & $\begin{array}{c} \textbf{99.16} \\ 99.10 \end{array}$ & $\begin{array}{c} \textbf{1e3} \\ 1.8e3 \end{array}$ & 8 & $\begin{array}{c} \textbf{75.35} \\ 75.09 \end{array}$ & $\begin{array}{c} \textbf{1.4e3} \\ 1.6e5 \end{array}$ & 10 & -- & -- & -- & -- & -- & --\\ \hline
    \end{tabular}}
\end{table}
\begin{table}
  \caption{Comparison of Bonsai with SpArSe for WM model \eqref{eq:workingmem}. Bonsai results taken from \citet{kumar2017resource}. The first row shows the highest accuracy model for WM $\leq 2$KB and the second row shows the highest accuracy model for WM, MS $\leq 2$KB. For MNIST, SpArSe is evaluated on the full ten-class dataset whereas Bonsai reports on a reduced two-class problem. SpArSe finds models with fewer parameters, less working memory, and higher accuracy in all cases. WM,MS reported in KB. Best performance highlighted in bold.}
  \vspace{3pt} 
  \label{table:working memory}
  \centering
  \resizebox{\linewidth}{!}{%
  \begin{tabular}{c|cccc|cccc|cccc|cccc}
        \hline
         & \multicolumn{4}{c}{MNIST} & \multicolumn{4}{c}{CIFAR10-binary} & \multicolumn{4}{c}{CUReT-binary} & \multicolumn{4}{c}{Chars4K-binary} \\
        &\rot{Acc} & \rot{WM} & \rot{MS} & \rot{GPUD} & \rot{Acc} & \rot{WM} & \rot{MS} & \rot{GPUD} & \rot{Acc} & \rot{WM} & \rot{MS} & \rot{GPUD} & \rot{Acc} & \rot{WM} & \rot{MS} & \rot{GPUD} \\ \hline
         SpArSe & \textbf{98.64} & 1.96 & 2.77 & 1 & \textbf{73.84} & \textbf{1.28} & \textbf{0.78} & 5 & \textbf{80.68} & 1.66 & 2.34 & 1 & \textbf{77.78} & \textbf{0.72} & \textbf{0.46} & 1 \\
         SpArSe & 96.49 & \textbf{1.33} & \textbf{1.44} & 1 & \textbf{73.84} & \textbf{1.28} &\textbf{ 0.78} & 5 & 79.97 & \textbf{1.43} & \textbf{1.69} & 1 & \textbf{77.78} & \textbf{0.72} & \textbf{0.46} & 1\\
         Bonsai & $94.38^*$ & $<2$ & $1.96$ & & $73.02$ & $<2$ & $1.98$ & &  -- & -- & -- & & $74.28$ & $<2$ & $2$ & \\ \hline
    \end{tabular}}
\vspace{-1em}
\end{table}

\begin{table}
  \caption{SpArSe versus Bonsai for WM model \eqref{eq:workingmem2}. See Table \ref{table:working memory} for details.}
  \label{table:working memory accurate}
  \centering
  \resizebox{\linewidth}{!}{%
  \begin{tabular}{c|cccc|cccc|cccc|cccc}
        \hline
         & \multicolumn{4}{c}{MNIST} & \multicolumn{4}{c}{CIFAR10-binary} & \multicolumn{4}{c}{CUReT-binary} & \multicolumn{4}{c}{Chars4K-binary} \\
        & \rot{Acc} & \rot{WM} & \rot{MS} & \rot{GPUD} & \rot{Acc} & \rot{WM} & \rot{MS} & \rot{GPUD} & \rot{Acc} & \rot{WM} & \rot{MS} & \rot{GPUD} & \rot{Acc} & \rot{WM} & \rot{MS} & \rot{GPUD} \\ \hline
         SpArSe & \textbf{97.03} & 1.38 & 15 & 1 & 70.41 & \textbf{0.91} & \textbf{1.98} & 18 & \textbf{73.22} & \textbf{1.9} & \textbf{0.14} & 2 & \textbf{76.83} & \textbf{0.39} & 20.12 & 1\\
         SpArSe & 95.76 & \textbf{0.62} & \textbf{1.76} & 2 & 70.41 & \textbf{0.91} & \textbf{1.98} & 18 & \textbf{73.22} & \textbf{1.9} & \textbf{0.14} & 2 & {74.87} & 1.64 & \textbf{0.16} & 3\\
          Bonsai & $94.38^*$ & $<2$ & $1.96$ & & \textbf{73.02} & $<2$ & \textbf{1.98} & & -- & -- & -- & & $74.71$ & $<2$ & $2$ & \\ \hline
    \end{tabular}}
\end{table}

\section{Conclusion}
Although MCUs are the most widely deployed computing platform, they have been largely ignored by ML researchers. 
This paper makes the case for targeting MCUs for deployment of ML, enabling future IoT products and usecases.
We demonstrate that, contrary to previous assertions, it is in fact possible to design CNNs for MCUs with as little as 2KB RAM.
SpArSe optimizes CNNs for the multiple constraints of MCU hardware platforms, finding models that are both smaller and more accurate than previous SOTA non-CNN models across a range of standard datasets.

\newpage

\bibliographystyle{plainnat}

\newpage
\begin{appendices}
\section{Pruning algorithm details}
Pruning can be expressed as
\begin{align}
    \label{eq:pruning-pruning}
    \bar{\omega} = \argmin_{\left(\sum_{G \in \mathscr{G}} \mathbbm{1}\left[\norm{\omega_{G}}_2 > 0 \right]\right) \leq s} \func{\mathscr{L}}{\braces{\alpha,\vartheta,\omega}}
\end{align}
where $\mathscr{L}(\cdot)$ denotes the loss function for the appropriate task, e.g. cross-entropy for classification, $\mathscr{G}$ denotes the set of disjoint groups covering the indices of each entry in $\omega$, $\omega_G$ denotes a particular group of weights, and $\mathbbm{1}\left[\cdot\right]$ denotes the indicator function. When $\vert G \vert = 1 \forall G \in \mathscr{G}$, \eqref{eq:pruning-pruning} is referred to as unstructured pruning. On other other hand, structured pruning arises when $\mathscr{G}$ is chosen to group related elements of $\omega$, i.e. the weights corresponding to a given feature map.

An alternative to \eqref{eq:pruning-pruning} is to cast pruning as Bayesian inference with priors that promote sparse solutions~\citep{tipping2001sparse}. 
One such algorithm for unstructured pruning is sparse variational dropout (SpVD) \citep{molchanov2017variational}. 
The prior over $\omega$ is assumed to factor over the elements of $\omega$, with $\func{p}{\vert \omega_{ij} \vert} \propto \inv{ \vert \omega_{ij} \vert}$. 
Given a dataset $\mathscr{D}$, the goal of Bayesian inference is to then compute the posterior $\func{p}{\omega | \mathscr{D}}$. 
SpVD employs variational inference (VI) \citep{jordan1999introduction} to approximate the posterior by a parametrized distribution $\func{q_\phi}{\omega}$, whose parameters $\phi$ are chosen to minimize $\func{D_{KL}}{\func{q_\phi}{\omega} || \func{p}{\omega | \mathscr{D}}}$. 
The distribution $\func{q_\phi}{\omega}$ is assumed to factor over the elements of $\omega$ and $\func{q_\phi}{\omega_{ij}} = \func{\mathsf{N}}{\mu_{ij},\beta_{ij} \mu_{ij}^2}$, where $\phi = \braces{\mu,\beta}$. 
Techniques for scalable VI are employed to estimate $\phi$ \citep{kingma2013auto,kingma2015variational}. 
Upon convergence, the estimate of $\bar{\omega}_{ij}$ becomes $\bar{\omega}_{ij} = \mu_{ij} \odot \mathbbm{1}\left[\beta_{ij} \leq \tau_l\right]$,
where $\tau_{l}$ is a layer-specific threshold and $\omega_{ij}$ resides in network layer $l$. 
Note that $\phi$ contains all of the information about both the network weight values as well as which weights can be masked to $0$. 
One of the side-effects of the choice of prior in SpVD is that the VI objective decomposes into a sum of a data-dependant term and a term which only depends on the prior, leading to the interpretation of VI as regularized training. 
Although there is no constant in front of the prior term, it can be beneficial to scale it by $\gamma$. 
Depending on the dataset, \citet{molchanov2017variational} keep $\gamma$ at $0$ for $N_1$ epochs, which is referred to as the pretraining phase, and then increase $\gamma$ to $\gamma_{N_2}$ over $N_2$ epochs \citep{sonderby2016train}. 
We include $\braces{\tau_l}_{l=1}^L$, $N_1$, $N_2$, and $\gamma_{N_2}$ in $\Omega$.

The structured pruning extension of SpVD is called Bayesian Compression (BC) \citep{louizos2017bayesian}, which assumes a hierarchical prior on $\omega$ that ties weights in the same group to each other: $\omega | z \sim \prod_{G \in \mathscr{G}} \prod_{(ij) \in G} \mathsf{N}(\omega_{ij}; 0,z_G^2)$. 
Inference for this prior proceeds in much the same way as SpVD and, upon convergence, entire groups of weights can be pruned away. 

\section{Search space details}
The search space considered in this work is described in Table \ref{table:search space}.

\begin{longtable}{|p{0.3\columnwidth}|p{0.2\columnwidth}|p{0.5\columnwidth}|}
\caption{Search space details. For discrete variables, ranges are listed in format [lowerbound:increment:upperbound].}\\
  Name & Range & Description \\ \hline
  downsample-input-in-depth & True/False & If True, max pool the input across the 3rd dimension \\ \hdashline
  downsample-input & True/False & If True, max pool the input in spatial dimensions \\ \hdashline
  input-downsampling-rate & $[2:1:4]$ & Active only if downsample-input = True. The amount by which to downsample the input.\\ \hdashline
  zero-regularization-epochs & [5:1:30] & Number of epochs for which VI inference is performed before the effect of the sparsity promoting prior is introduced. \\ \hdashline
  annealing-epochs & [15:1:25] & Only active if pretraining=False. Number of epochs over which the coefficient in front of the regularization term in the VI objective is annealed from $0$ to its final value \\ \hdashline
  $\alpha$ & [1e-2:1e-2:1] & Final value for the coefficient of the regularization term in the VI objective \\ \hdashline
  pretraining & True/False & Only active if pretraining=False. If True, pretrain the CNN before pruning \\ \hdashline
  batch-norm & True/False & Only used for random weight pruning experiments. If True, apply batch-normalization to the output of each layer\\ \hdashline
  num-conv-blocks & [1:1:2] & Number of convolution blocks in the CNN, where each block consists of a series of convolutional layers. The output of each block is downsampled through max pooling \\ \hdashline
  num-fc-layers & [0:1:1] & Number of FC layers in the main branch following the convolution blocks \\ \hdashline
  pruning-thresholds-block-$k$-layer-$l$ & [-6:1e-1:3] & Thresholds for pruning weights in block $k$ layer $l$ \\ \hdashline
  total-fc-layer-weights & [1:1:800]e3 & Number of weights in the FC layers comprising the main, left, and right branches \\ \hdashline
  weight-fraction-main-branch & [0:1] & Percentage of total-fc-layer-weights that go into the FC layer in the main branch\\ \hdashline
  num-conv-layers-block-$k$ & [1:1:3] & Number of convolutional layers in block $k$ \\ \hdashline
  layer-type-block-$k$-layer-$l$ & [Conv2D, DownsampledConv2D, SeparableConv2D] & Layer type for convolutional block $k$ layer $l$ \\ \hdashline
  kernel-size-block-$k$-layer-$l$ & [2:1:5] & Convolutional kernel size of block $k$ layer $l$ \\ \hdashline
  num-filters-block-$k$-layer-$l$ & [1:1:100] & Number of output feature maps for block $k$ layer $l$ \\ \hdashline
  downsample-block-$k$-layer-$l$ & [0:0.5] & Active only if layer-type-block-$k$-layer-$l$=DownsampledConv2D. If True, the input feature maps are first passed through a $1\times 1 \times (downsample-block-k-layer-l \times num-filters-block-k-layer-(l-1))$ convolutional layer \\ \hdashline
  left-branch & True/False & If True, a branch is added to the feed-forward architecture. The branch takes the output of the first convolution block, sends it to an FC layer, sends the result to a merge operation, whose output is sent to a final FC layer \\ \hdashline
  right-branch& True/False & If True, a branch is added to the feed-forward architecture. The branch takes the input to the first convolution block, sends it to an FC layer, sends the result to a merge operation, whose output is sent to a final FC layer \\\hdashline
  weight-fraction-left-branch & [0.01:1] & Active only if left-branch=True. Percentage of total-fc-layer-weights that go into the left branch FC layer \\\hdashline
  weight-fraction-left-branch & [0.01:1] & Active only if right-branch=True. Percentage of total-fc-layer-weights that go into the right branch FC layer \\\hdashline
  merge-type & Sum/Concatenate & Active only if at least one of left-branch or right-branch are True. How the main, left, and right branches are to be combined\\ \hline
  \label{table:search space}
\end{longtable}
 
\section{Morphism detals}
In the present work, a configuration $\Omega^n$ is considered a morph of $\Omega^r$ if $\Omega^n$ is generated by applying one or more of the operations listed in Table \ref{table:morphs} to $\Omega^r$. These morphs are used to generated random samples for the Thompson sampling step in Section \ref{sec:MOBO}. Each sample $\Omega^n$ is generating by randomly choosing one or more of the morphs from Table $\ref{table:morphs}$ and applying them to a randomly chosen $\Omega^r \in \braces{\Omega^r}_{r=1}^{n-1}$. This procedure ensures that each configuration proposal is relatively close to a reference configuration. We then use the fact that $\Omega^n$ is closely related to $\Omega^r$ during the pruning process by letting $\phi^n$ inherit information from $\phi^r$, where $\phi^n$ denotes the parameters of the approximated weight posterior for configuration $\Omega^n$. The inheritance process proceeds by first checking for identical nodes between $\Omega^r$ and $\Omega^n$ and then copies the corresponding elements of $\phi^r$ into $\phi^n$ for those nodes. The nodes which participate in this step are the nodes which were not influenced during the morphing process. For the remaining nodes, if corresponding nodes in $\Omega^n$ and $\Omega^r$ have the same operation type, we copy as many of the corresponding elements of $\phi^r$ into $\phi^n$ as possible. For example, if the first layer of $\Omega^n$ is a $3\times 3 \times 50$ convolution and the first layer of $\Omega^r$ is a $3\times 3\times 30$ convolution, we copy the elements of $\phi^r$ corresponding to the first convolution layer into the elements of $\phi^n$ corresponding to the first $30$ feature maps of the first convolution layer. Upon completion of the inheritance process, most of the elements of $\phi^n$ are inherited from $\phi^r$, and the remaining elements are learned from the training data. Unlike \citet{elsken2018efficient}, we do not restrict the training process to just the elements of $phi^n$ which were not inherited, but instead update all of the elements of $\phi^n$ during learning.
 
\begin{table}
  \caption{Allowable morphs. Random sampling is always performed under a uniform distribution.}
  \label{table:morphs}
  \centering
  \resizebox{\linewidth}{!}{%
  \begin{tabular}{|p{0.3\columnwidth} | p{0.7\columnwidth}|}
  \hline
  Morph & Description \\ \hline
  num-fc-layers & Change the number of FC layers in the main branch by $\pm 1$ \\ \hdashline
  num-conv-blocks & Change the number of convolution blocks by $\pm 1$. If the number of convolution blocks is increased, set the number of convolution layers in the new block to $1$ \\ \hdashline
  layer-type & Change the layer type for a randomly chosen convolution layer \\ \hdashline 
  num-conv-filters & Change the number of output feature maps in a randomly chosen convolution layer \\ \hdashline 
  kernel-size & Change the kernel size for a randomly chosen convolution filter \\ \hdashline 
  downsampling-rate & Randomly choose a convolution layer that has type DownsampledConv2D and randomly sample its downsampling rate \\ \hdashline batch-norm & Switch the state of the batch-norm parameter \\ \hdashline
  residual-connections & Switch the state of the residual connections parameter \\ \hdashline
  left-branch & Switch the state of the left-branch parameter. If the new state is True, set weight-fraction-left-branch=0.05 \\ \hdashline
  right-branch & Switch the state of the right-branch parameter. If the new state is True, set weight-fraction-right-branch=0.05 \\ \hdashline
  total-fc-layer-weights & Change the number of total FC layer weights by $\pm$5e3 \\ \hdashline
  merge-type & Switch the state of the merge-type parameter \\ \hdashline
  threshold & Change the value of a randomly chosen pruning threshold by 0.5 \\ \hdashline
  weight-fraction & For each one weight-fraction-main-branch, weight-fraction-left-branch, weight-fraction-right-branch, perturb each active parameter by 5e-2 \\ \hdashline
  $\alpha$ & Change $\alpha$ by $\pm 0.1$ \\ \hdashline 
  num-conv-layers & Change the number of convolution layers in a randomly chosen convolution block by $\pm 1$ \\ \hline
  \end{tabular}}
\end{table}
\section{Visualization of discovered CNNs}
Fig. \ref{fig:winning_cnn_arch} shows the architectures which dominated the competing methods in Table \ref{table:dominating}.

\section{Extended results on interaction of pruning and architecture}
\begin{figure*}
    \centering
    \begin{subfigure}[t]{0.45\textwidth}
        \centering
        \includegraphics[width=\textwidth]{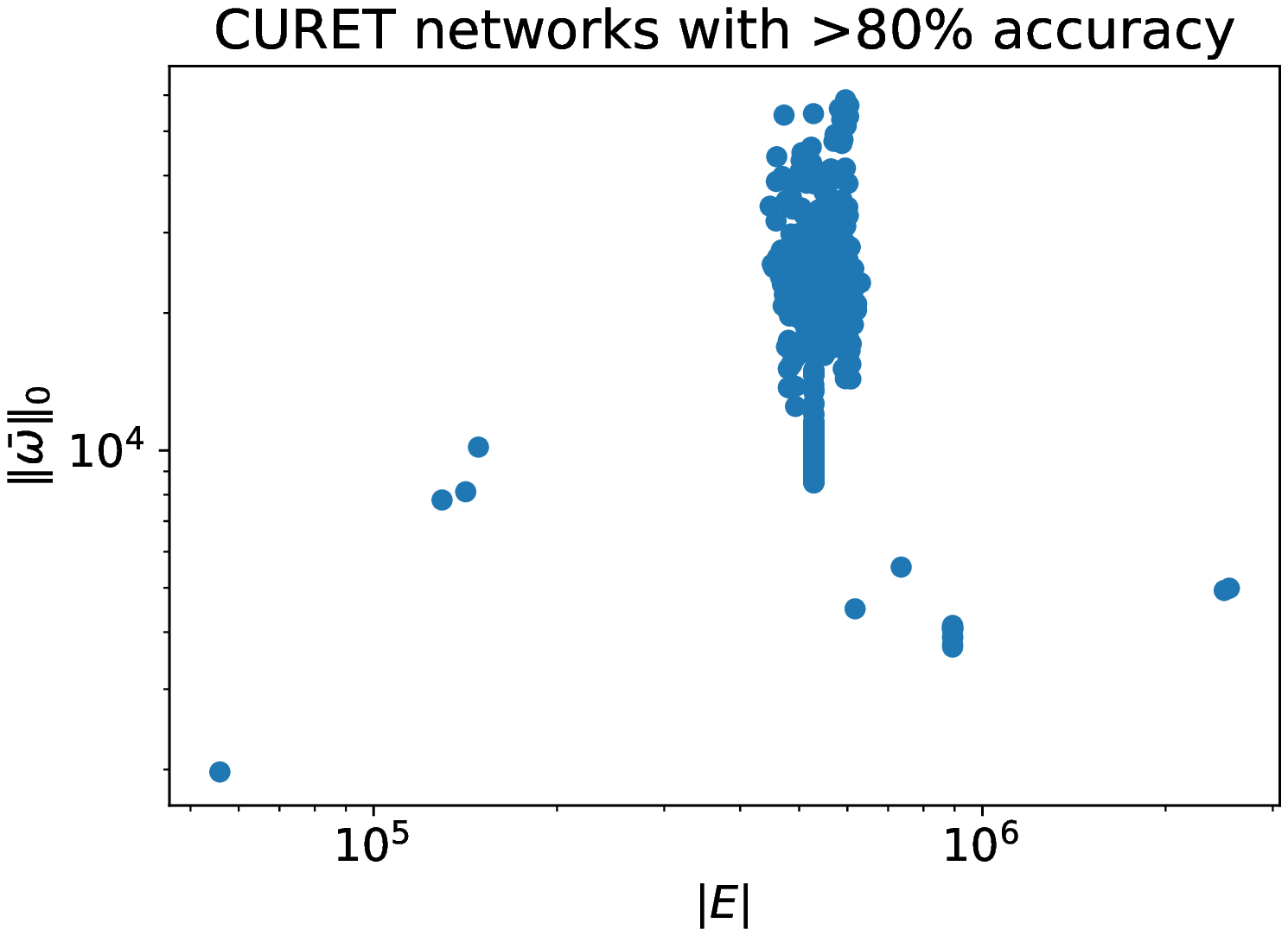}
        \caption{}
    \end{subfigure}%
    ~ 
    \begin{subfigure}[t]{0.45\textwidth}
        \centering
        \includegraphics[width=\textwidth]{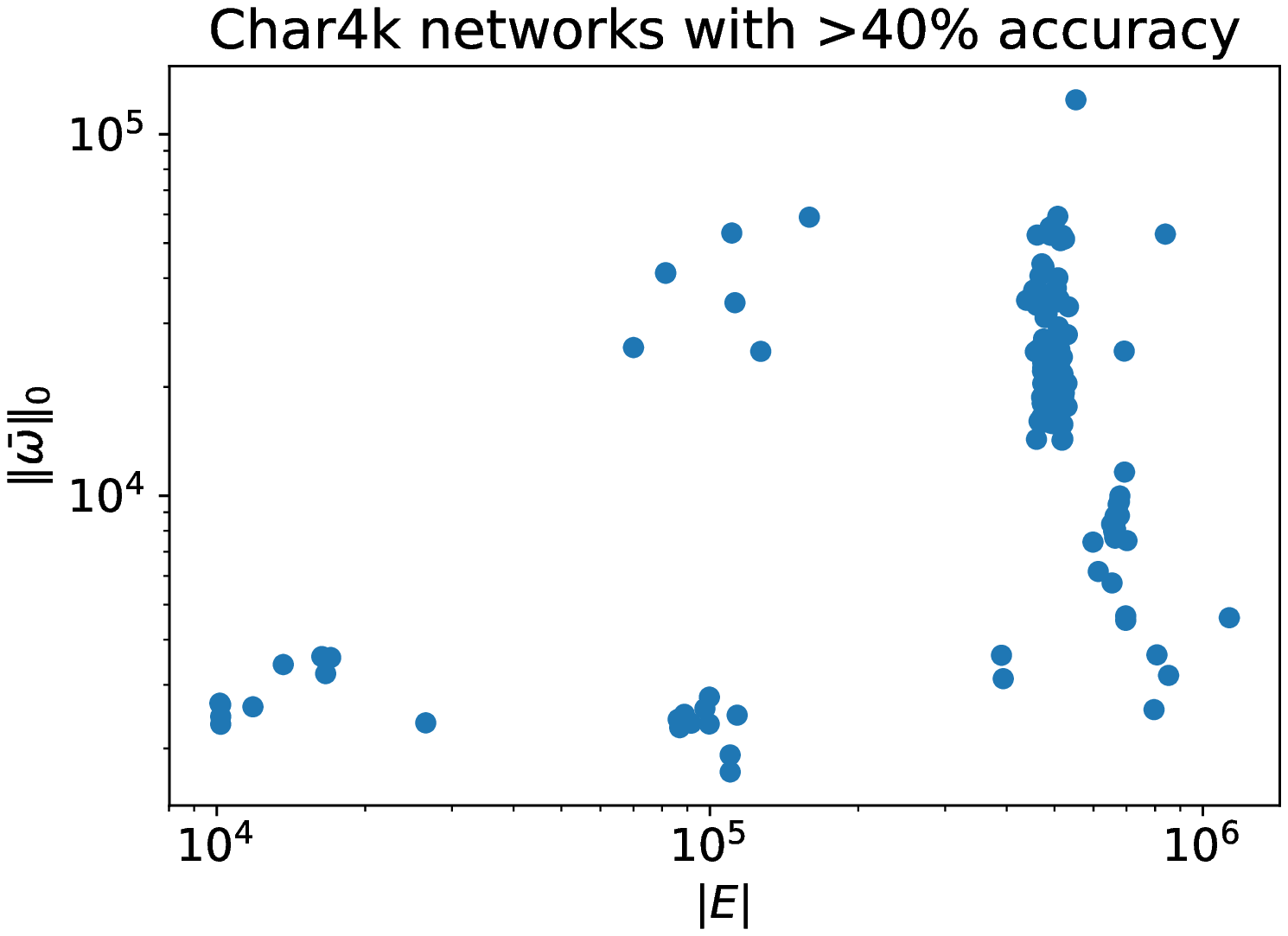}
        \caption{}
    \end{subfigure}
    \caption{Scatter plots of $\vert V \vert$ versus $\norm{\bar{\omega}}_0$ for the best performing configurations.}
    \label{fig:scatter extended}
\end{figure*}

Fig. \ref{fig:scatter extended} shows the interaction of pruning with architecture for the Chars4k and CUReT dataset experiments in Section \ref{sec:parameter minimization}.
MNIST and CIFAR10-binary are given in Fig.~\ref{fig:scatter}.

\section{Evolution of winning CNNs}
\begin{figure*}
    \centering
    \centering
    \begin{subfigure}[t]{0.45\textwidth}
        \centering
        \includegraphics[width=\textwidth]{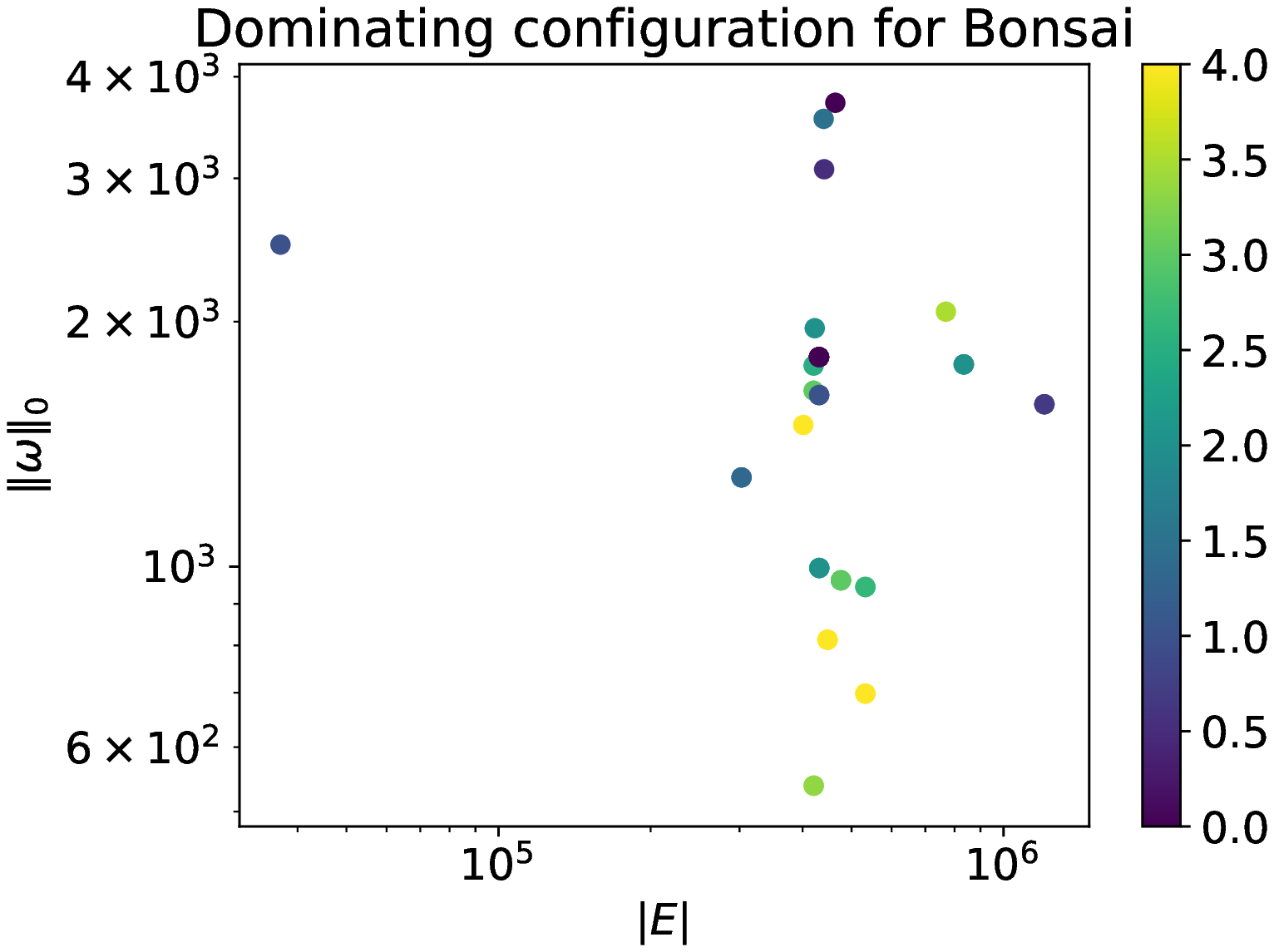}
        \caption{}
    \end{subfigure}
    ~
    \begin{subfigure}[t]{0.45\textwidth}
        \centering
        \includegraphics[width=\textwidth]{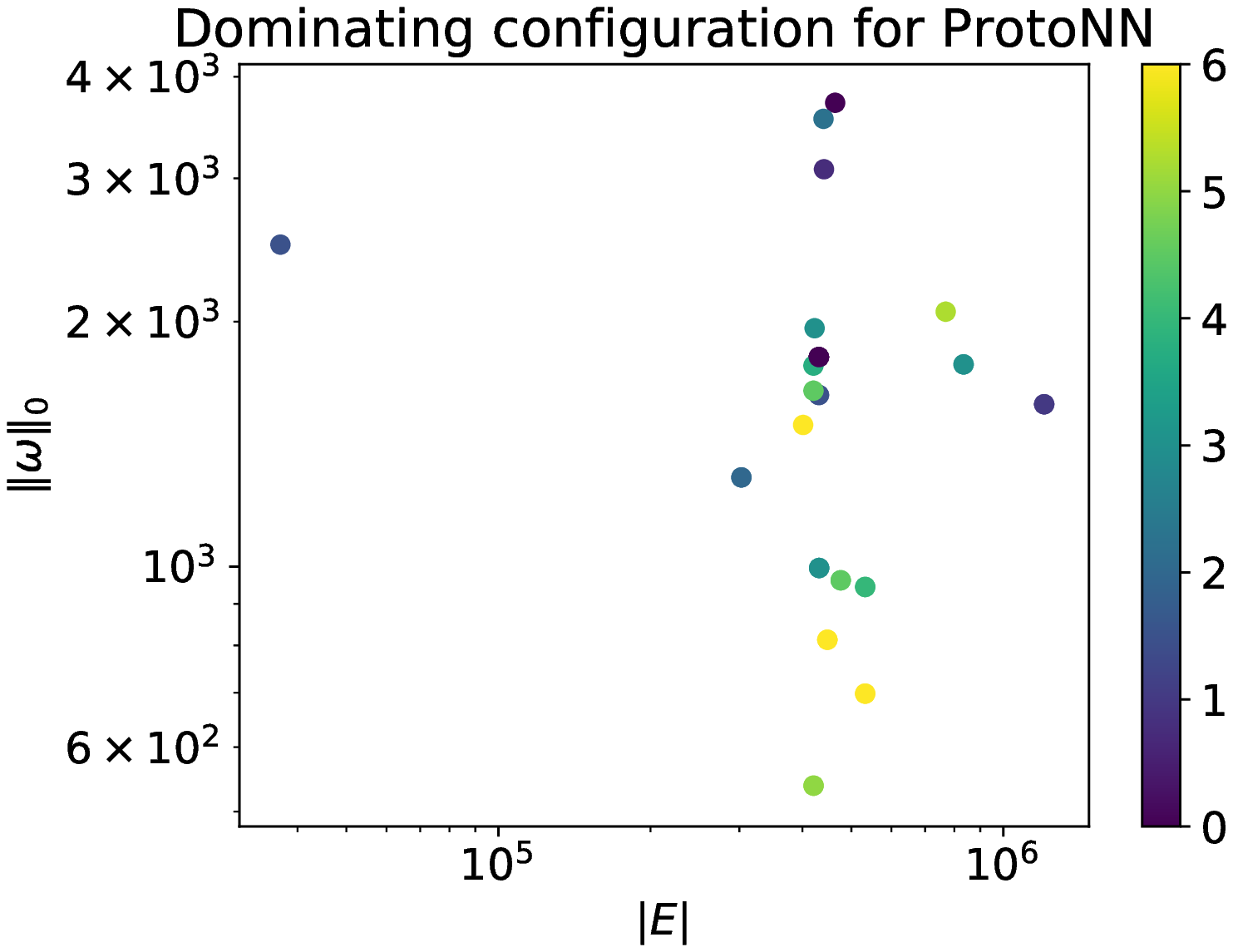}
        \caption{}
    \end{subfigure}
    ~
    \begin{subfigure}[t]{0.45\textwidth}
        \centering
        \includegraphics[width=\textwidth]{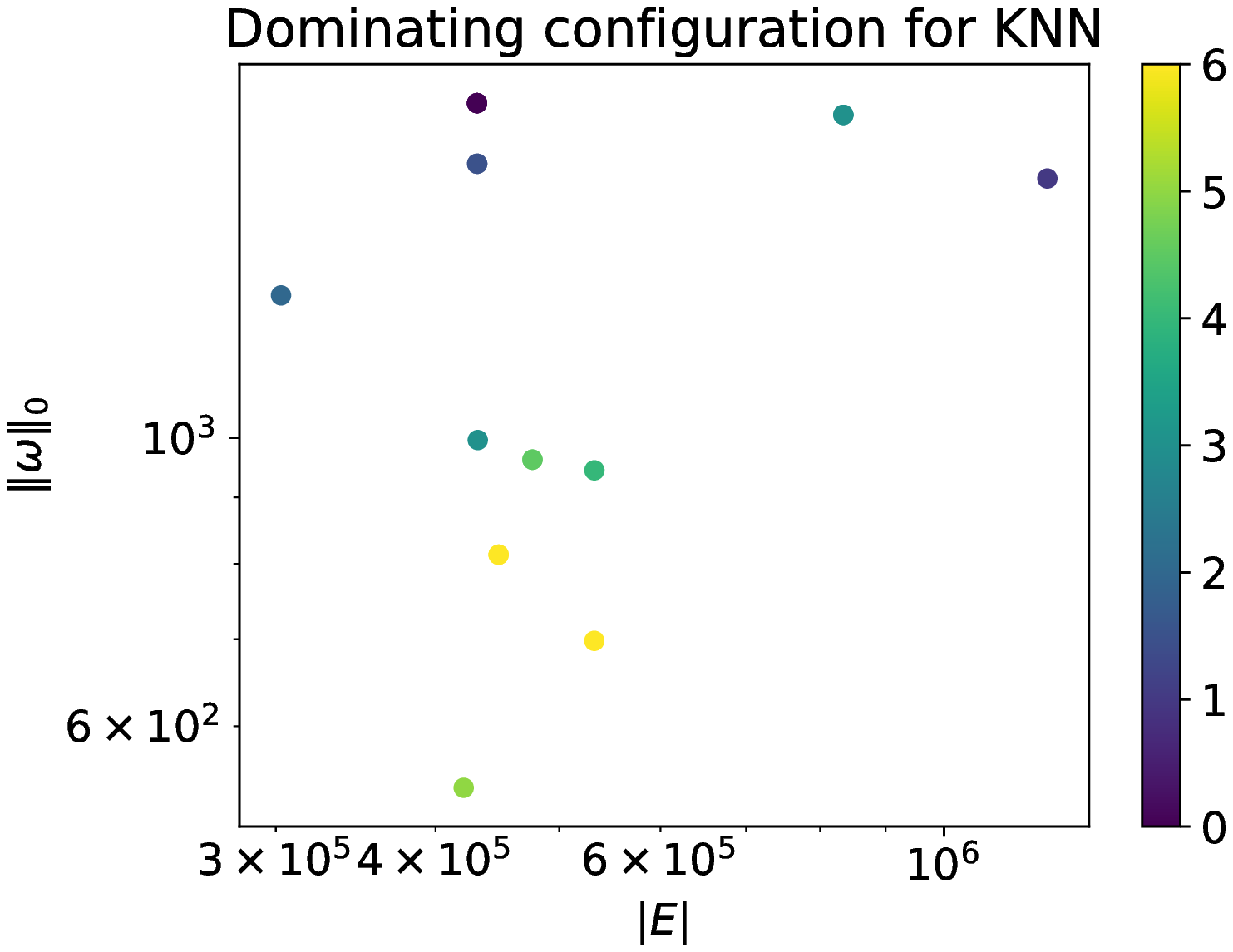}
        \caption{}
    \end{subfigure}
    ~
    \begin{subfigure}[t]{0.45\textwidth}
        \centering
        \includegraphics[width=\textwidth]{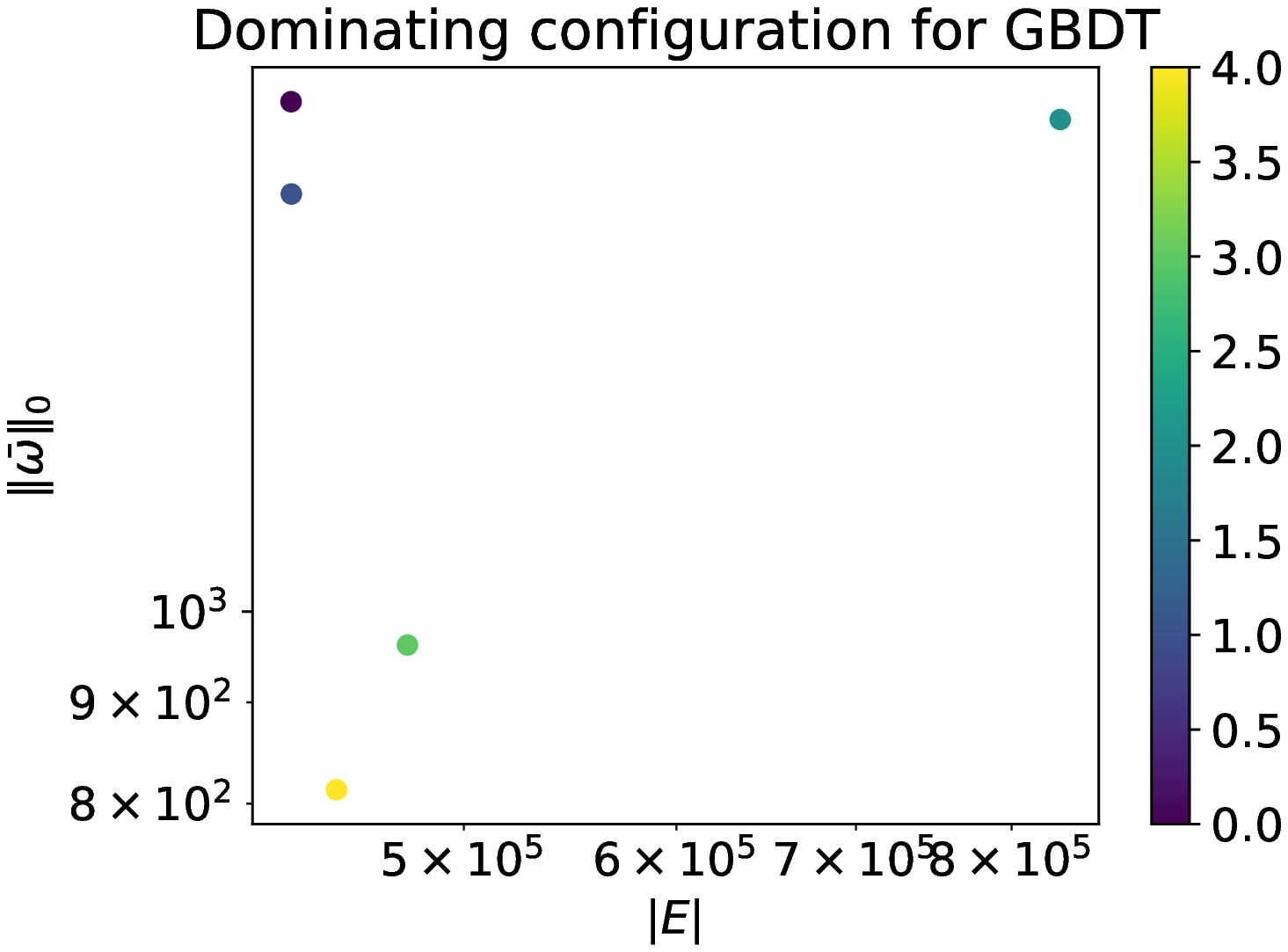}
        \caption{}
    \end{subfigure}
    ~
    \begin{subfigure}[t]{0.45\textwidth}
        \centering
        \includegraphics[width=\textwidth]{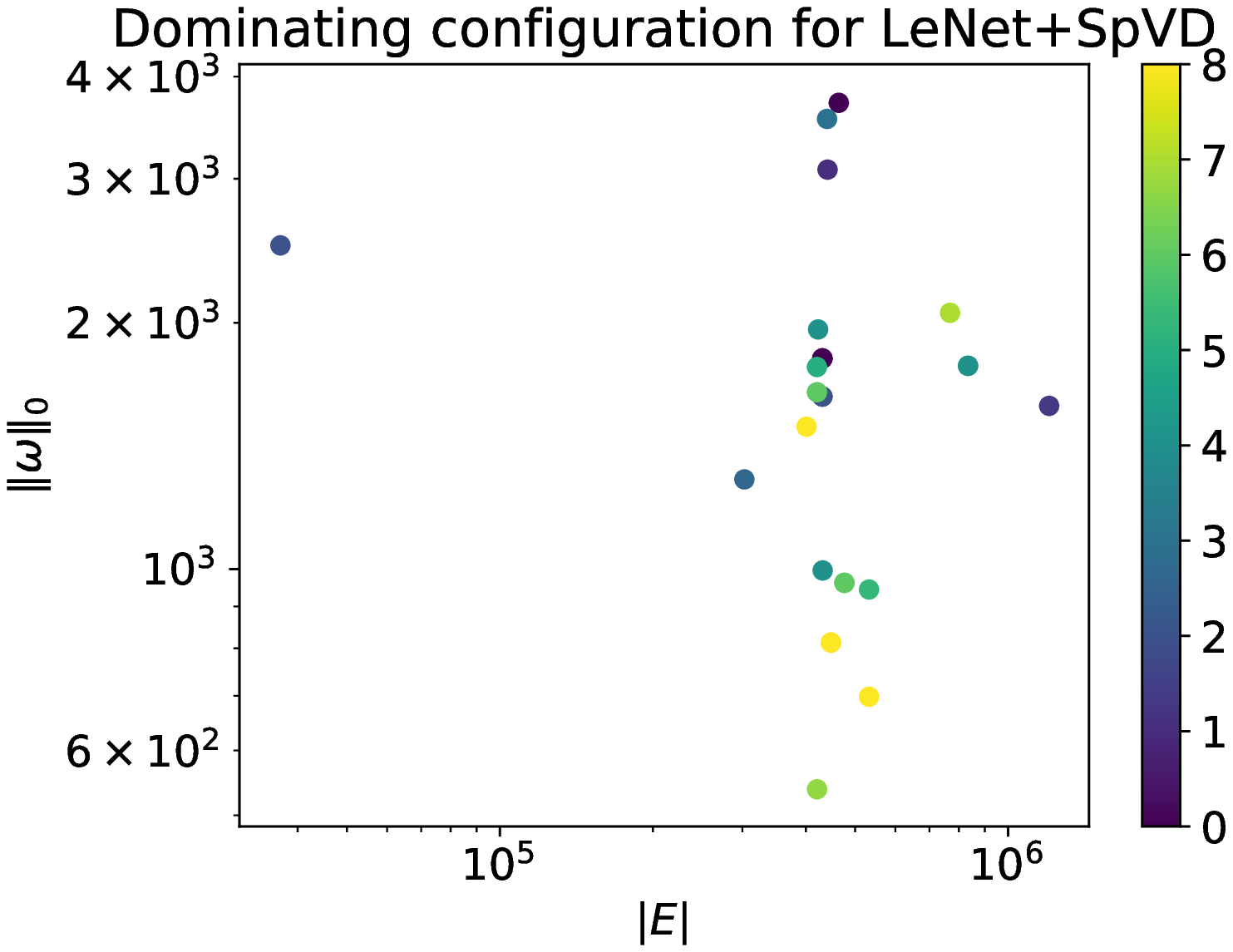}
        \caption{}
    \end{subfigure}
    ~
    \begin{subfigure}[t]{0.45\textwidth}
        \centering
        \includegraphics[width=\textwidth]{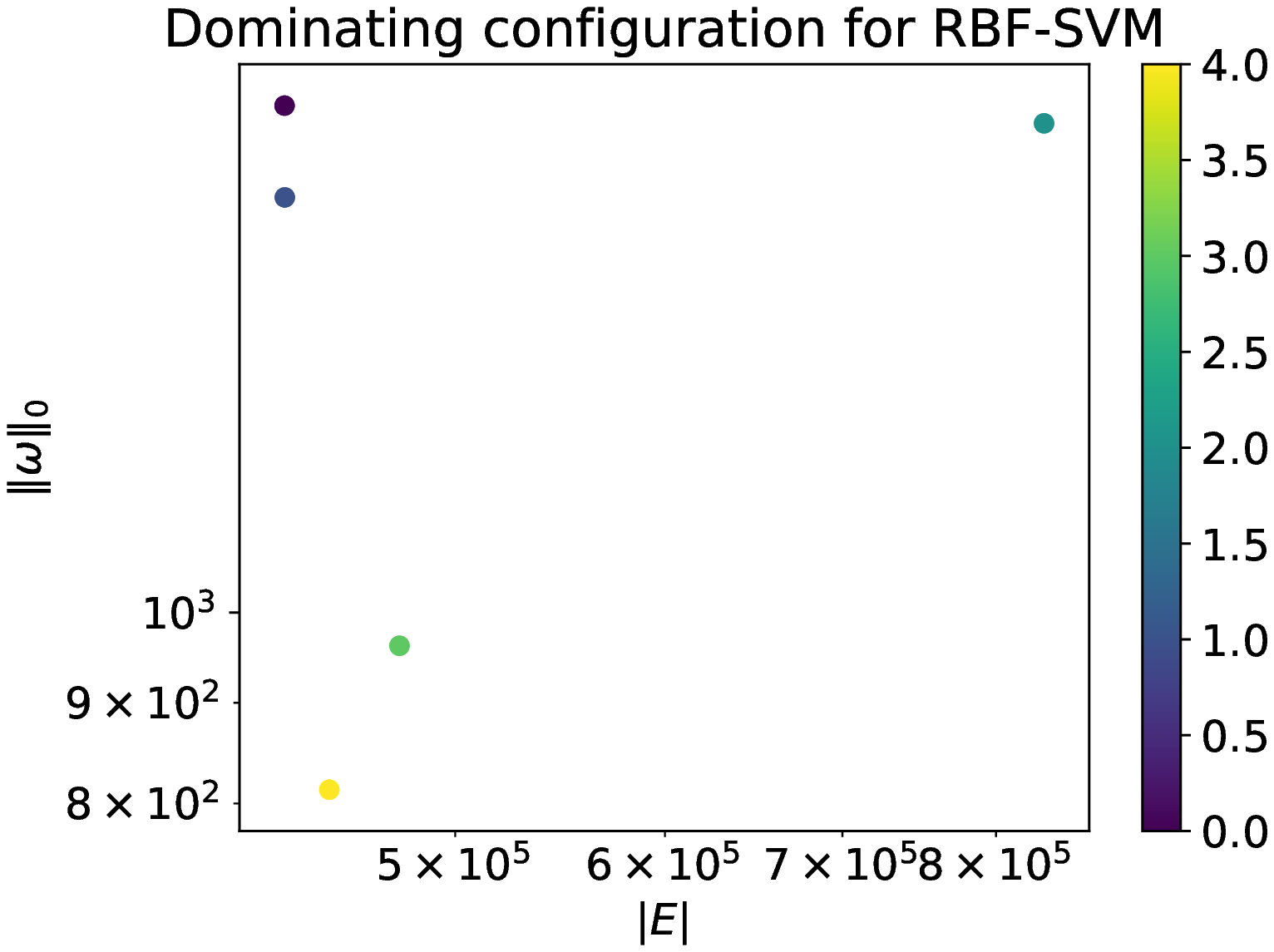}
        \caption{}
    \end{subfigure}
    \caption{MNIST: Evolution of dominating configuration. Lighter colored samples indicate configurations which were sampled later in the optimization process.}
    \label{fig:evolution}
\end{figure*}

The evolution of the CNN architectures which ended up dominating the competing methods on the MNIST dataset in Table \ref{table:dominating} is shown in Fig. \ref{fig:evolution}.

\section{Visualization of winning CNNs}
\begin{figure*}
    \centering
    \begin{subfigure}[t]{0.45\textwidth}
        \centering
        \includegraphics[width=\textwidth]{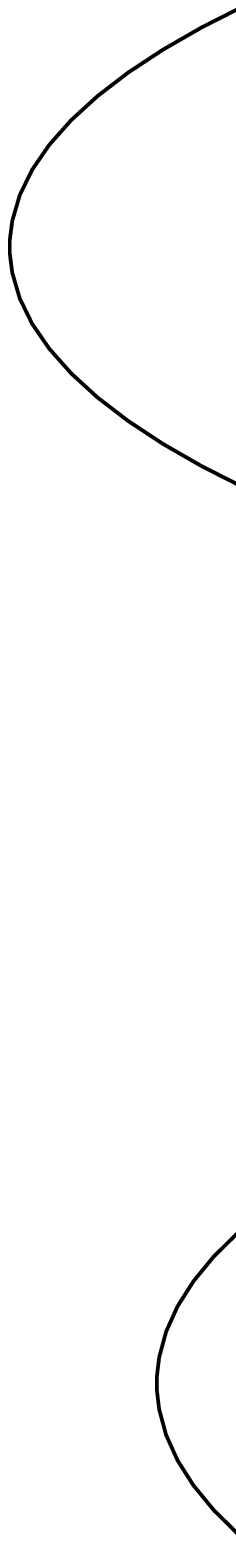}
        \caption{Winner against Bonsai on MNIST}
    \end{subfigure}
    ~
    \begin{subfigure}[t]{0.45\textwidth}
        \centering
        \includegraphics[width=\textwidth]{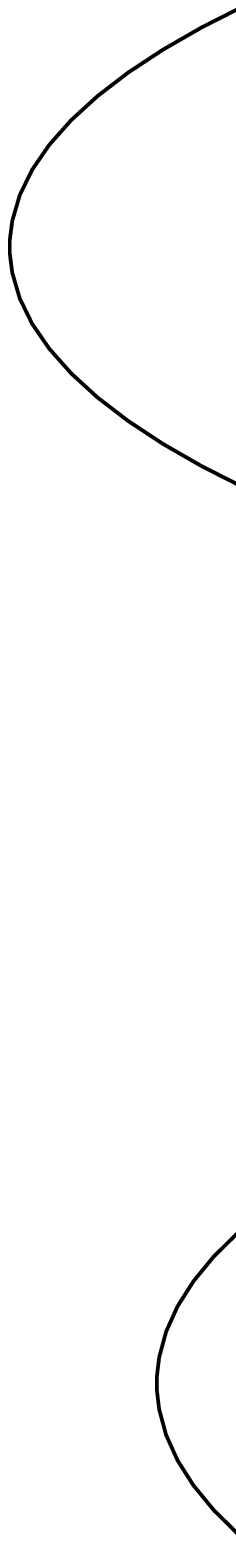}
        \caption{Winner against ProtoNN on MNIST}
    \end{subfigure}
    ~
    \begin{subfigure}[t]{0.45\textwidth}
        \centering
        \includegraphics[width=\textwidth]{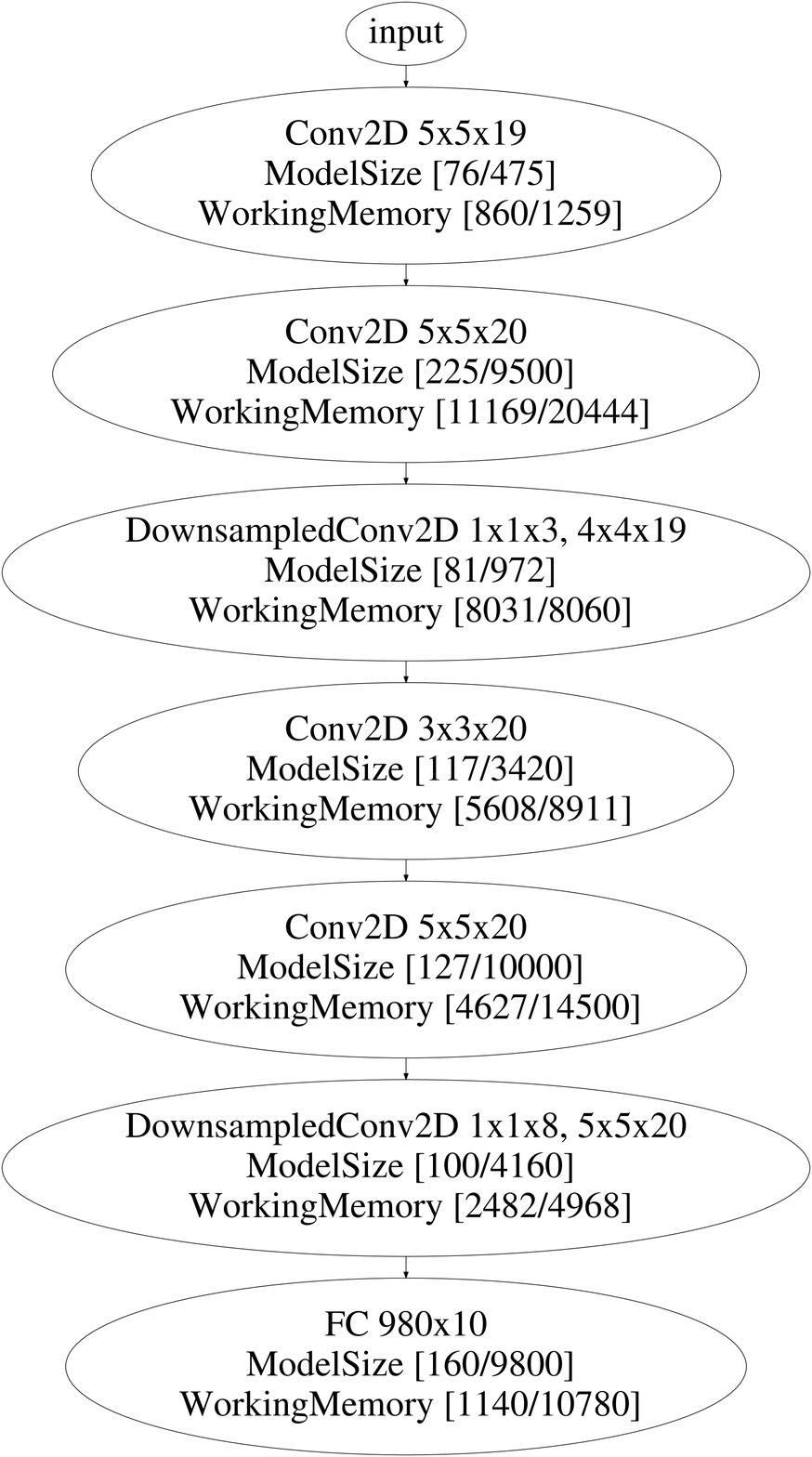}
        \caption{Winner against LeNet+SpVD on MNIST}
    \end{subfigure}
    ~
    \begin{subfigure}[t]{0.45\textwidth}
        \centering
        \includegraphics[width=\textwidth]{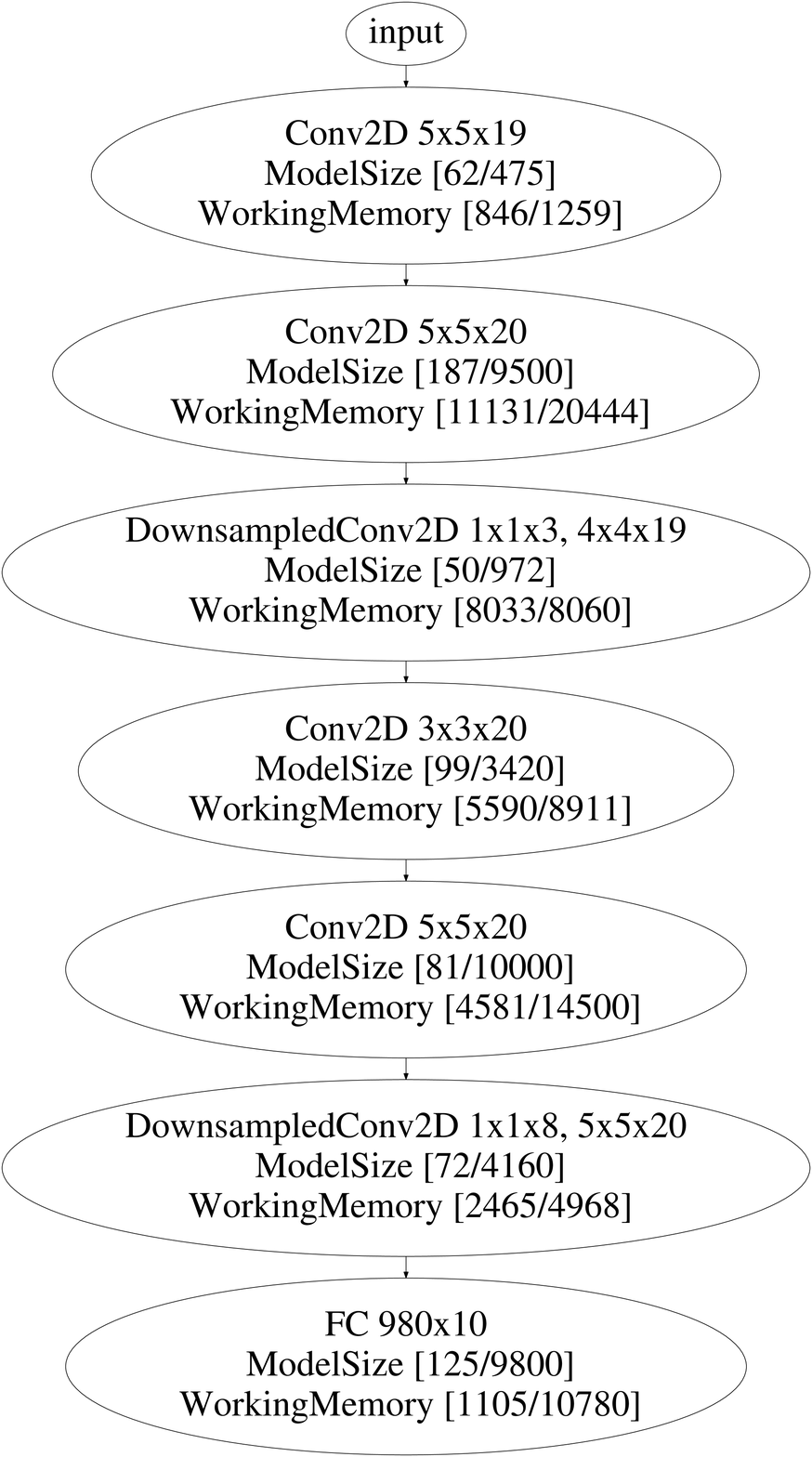}
        \caption{Winner against GBDT on MNIST}
    \end{subfigure}
    \caption{Visualization of winning CNNs on MNIST classification in Table \ref{table:dominating}. Working memory is reported for the model in \eqref{eq:workingmem}. The dominating configuration against KNN is the same as that for ProtoNN.  The dominating configuration against RBF-SVM is the same as that for Bonsai.}
    \label{fig:winning_cnn_arch}
\end{figure*}

Fig. \ref{fig:winning_cnn_arch} shows the architectures which dominating the competing methods on the task of classifying MNIST with the minimum number of parameters (i.e. Section \ref{sec:parameter minimization}).

\end{appendices}

\end{document}